\documentclass[10pt,twocolumn,letterpaper]{article}

\usepackage{lipsum}
\usepackage{iccv}
\usepackage{times}
\usepackage{epsfig}
\usepackage{graphicx}
\usepackage{amsmath}
\usepackage{amssymb}
\usepackage{booktabs}
\usepackage{ulem}
\usepackage{bbding}
\usepackage{float}
\usepackage{makecell}
\usepackage{multirow}
\usepackage{ulem} 
\usepackage{color}
\usepackage[accsupp]{axessibility} 
\usepackage[colorlinks,linkcolor=blue]{hyperref}

\newcommand\blfootnote[1]{%
	\begingroup
	\renewcommand\thefootnote{}\footnote{#1}%
	\addtocounter{footnote}{-1}%
	\endgroup
}

\iccvfinalcopy % *** Uncomment this line for the final submission

\def\iccvPaperID{9149} % *** Enter the ICCV Paper ID here

\ificcvfinal\fi

\begin{document}
	%\setlength{\textfloatsep}{6pt}
	
	%%%%%%%%% TITLE
	\title{FineDance: A  Fine-grained Choreography Dataset \\ for 3D Full Body Dance Generation}% Hand-aware

	\author{{Ronghui Li}$^{1*}$, {Junfan Zhao}$^{1*}$
		,Yachao Zhang$^1$, \\Mingyang Su$^1$,
		Zeping Ren$^1$,Han Zhang$^2$,Yansong Tang$^1$, Xiu Li$^{1\dagger}$\\
		$^1$Tsinghua Shenzhen International Graduate School, Tsinghua University\\
		%School of Artificial Intelligence, Optics and Electronices(iOPEN),\\
		$^2$Northwestern Polytechnical University\\
		\small{\{lrh22, yf-zhao21\}@mails.tsinghua.edu.cn, \{yachaozhang, tang.yansong, li.xiu\}@sz.tsinghua.edu.cn}
	% \and
}

% \maketitle
\twocolumn[{ 
	\maketitle
	\begin{figure}[H]
		\hsize=\textwidth  
		\centering
		\includegraphics[width=\textwidth]{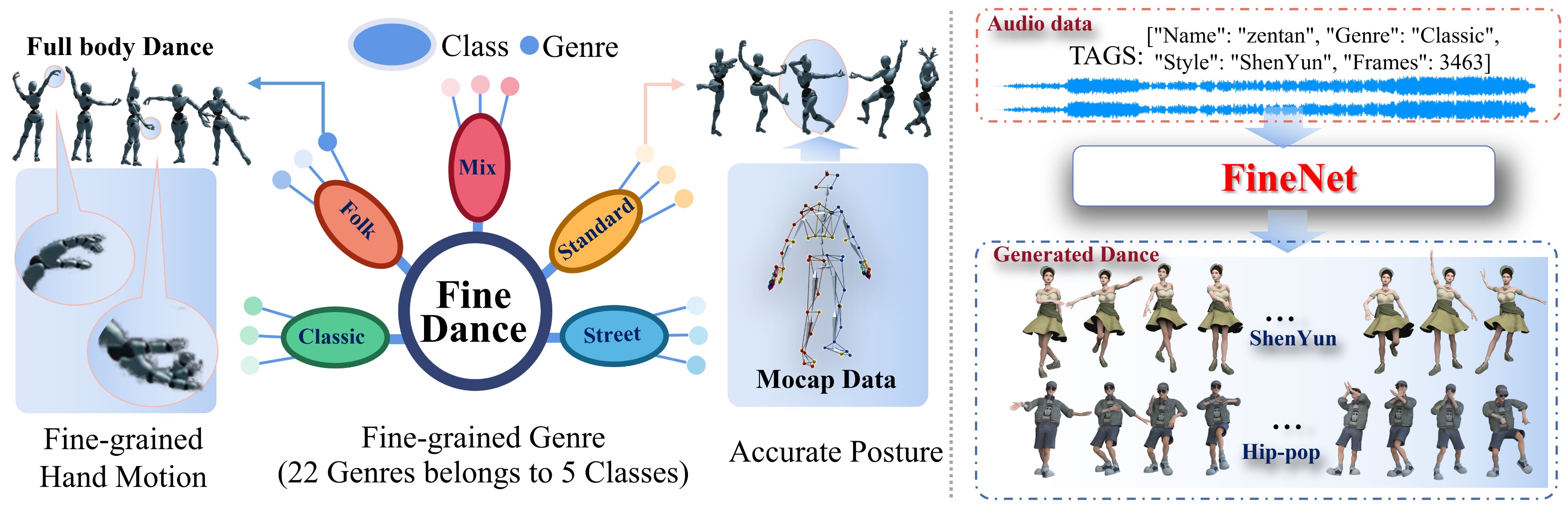}%images/
		\caption{A conceptual overview. We release a large-scale professional 3D motion capture dance dataset \textbf{FineDance}, and propose a choreography network \textbf{FineNet}. Training with FineDance, FineNet can generate multi-genre dances with expressive hand movements.}%The FineDance and FineNet.
		\label{fig:framework}
	\end{figure}
}]


%%%%%%%%% ABSTRACT
\begin{abstract}
	Generating full-body and multi-genre dance sequences from given music is a challenging task, due to the limitations of existing datasets and the inherent complexity of the fine-grained hand motion and dance genres. 
	\blfootnote{
		$*$ equal contribution, 
		$\dagger$ corresponding author}
	To address these problems, we propose FineDance, which contains 14.6 hours of music-dance paired data, with fine-grained hand motions, fine-grained genres (22 dance genres), and accurate posture. To the best of our knowledge, FineDance is the largest music-dance paired dataset with the most dance genres. 
	Additionally, to address monotonous and unnatural hand movements existing in previous methods, we propose a full-body dance generation network, which utilizes  the diverse generation capabilities of the diffusion model to solve monotonous problems, and use expert nets to solve unreal problems.
	To further enhance the genre-matching and long-term stability of generated dances, we propose a Genre$\&$Coherent aware Retrieval Module. 
	Besides, we propose a novel metric named Genre Matching Score to evaluate the genre-matching degree between dance and music. 
	Quantitative and qualitative experiments demonstrate the quality of FineDance, and the state-of-the-art performance of FineNet. The FineDance Dataset and more qualitative samples
	can be found at \href{https://li-ronghui.github.io/finedance}{website}.
	
\end{abstract}
% \footnote{text for footnote}
%%%%%%%%% BODY TEXT
\section{Introduction}

%Music and dance are two enduring art forms that can embody human emotions, which are widely used in concerts, movies, games and other industries in modern society\cite{hanna1987dance}.
%However, acquiring high-quality 3D dance animation is very expensive, often requiring experienced dancers, engineers, and expensive motion-capture equipment to capture the 3D dance\cite{ChoreMaster}. 
%The correspondence between music and dance genres is a key aspect in assessing the quality of generated dance. However, there is currently no good method for calculating this metric, due to the different modalities of music and dance. To address this issue, we propose a genre matching score calculation network and introduce the GS metric to measure the level of correspondence between music and dance genres.

%Therefore, Using AI to generate 3D dance from music is becoming a hot research topic.
Music and dance are two enduring art forms that can express a wide range of human emotions, and they have become essential elements in modern entertainment industries such as concerts, movies, and games\cite{hanna1987dance}. However, creating high-quality 3D dance animations can be a costly and complex process that often involves skilled dancers, engineers, and expensive motion-capture equipment\cite{ChoreMaster}. 
As a result, there is growing interest in using artificial intelligence (AI) to generate 3D dance animations from music, which has become a rapidly developing research topic.

% 这里我准备分数据集和方法两方面写，这一段是原来的
Despite the wide range of research in this field
\cite{dancingtomusic,dancerevolution,aist++,danceformer,GrooveNet,Music2Dance,dancewithmelody,ChoreMaster, ye2020choreonet, tang2023flag3d,ma2023follow,zhang2021we,zhang2022wanderings}, enerating high-quality dances is  still limited by the existing datasets: 
(1) \textbf{Full-body expressiveness:} Existing dance datasets contain few hand movements, and only 1.5 hours of full body dance data is available. 
%The existing dance datasets either do not include hand movements, or only 1.5 hours of full-body dance data are available.
%The previous methods of motion generation, treating body and hand in the same way, get unnatural or monotonous hand motions, due to the body and hand in different feature spaces. The uncoordinated body and hand motions can destroy the expressiveness of the overall dance. 
The previous methods in motion generation, treating body and hand as the same, lead to unnatural or monotonous hand motions, because the body and hand are in different feature spaces. However, uncoordinated body and hand motions can destroy the expressiveness of the overall dance. 
(2) \textbf{Multi-genre:} Existing datasets contain a limited number of dance genres, so the generated dances are not sufficient to match various music styles. Previous methods struggle with limited coarse dance genres and have no suitable objective metric to measure the genre-matching degree between music and generated dances. 
% 第三点 long-term删掉了 看一下上面这句是这个意思吧？
% (3) Duration adaptability: The choreography algorithm should generate coherent and long-term dances based on the music of any duration.
% Despite the wide range of research in this field\cite{dancingtomusic,dancerevolution,aist++,danceformer,GrooveNet,Music2Dance,dancewithmelody,ChoreMaster, ye2020choreonet}, generating high-quality dances is still difficult due to the limitations of existing datasets and methods.

% 数据集介绍部分删掉了
%Existing available 3D dance datasets \cite{dancingtomusic,dancerevolution,aist++,GrooveNet,Music2Dance,dancewithmelody} are limited to a few selected genres and do not support multi-genre dance generation. For instance, AIST++ \cite{aist++} only includes street dance, and Dance with Melody \cite{dancewithmelody} contains just four genres, resulting in a total of 1.56 hours of dances. 
To address the limitations of existing datasets, we introduce a Fine-grained Choreography Dance dataset (FineDance).
It comprises over 14.6 hours of data collected from 346 paired songs and dances, was created by professional dancers and a motion capture system, which has accurate body and hand motions.
The fine-grained 22 dance genres of FineDance spanning traditional and modern styles, which make the genre-matching of generated dance sequences and given music become more challenging.
%with 22 dance genres, spanning traditional and modern styles. The fine-grained 22 dance genres make the genre-matching of generated dance sequences and given music become more challenging.
FineDance includes music, dance sequences, FilmBox (fbx) files, SMPL\cite{SMPL,SMPLX}, and multi-view videos.

Early music-driven dance synthesis methods \cite{arikan2002interactive,kim2003rhythmic, lucas2002motion, shiratori2006dancing, kim2006making, dancewithmelody, GrooveNet, zhang2022learning} often rely on motion graph-based algorithms where the dance fragments from a pre-existing music-dance database are stitched together to synthesize one dance. %that stitch together dance fragments from a pre-existing music-dance database. 
While such methods can synthesize long-term dances, they do not produce new dance fragments. Recently methods have employed generative networks such as VAE\cite{siyao2022bailando}, GAN\cite{Self_supervised_dance_synthesis}, Normalization Flow Network\cite{valle2021transflower}, Diffusion\cite{edge}. But they focus solely on body part, while neglecting hand movements, resulting in unnatural or monotonous hand motions even trained with well-annotated body and hand labels. Additionally, the generative-based methods are limited by the long-term modeling ability of the networks, making them difficult to generate long-term dance sequences. 

Therefore, we propose FineNet, a two-stage generative-synthesis network that addresses the limitations of previous dance generation methods. In the first stage, we propose a diffusion-based Full-body dance generation network (FDGN). The key of FDGN is to design two expert networks, which are dedicated to the generation of body and hand motions, and use a Refine Net to assemble them coordinately. In the second stage, we propose a Genre$\&$Coherence aware Retrieval Module (GCRM), which ensures the coherence of dance fragments and matches the genre between the music and the dances.
% 这里指标简化成一句话
Based on the suitable dance fragments retrieved by GCRM, we can produce genre-matching and long-term dances. A conceptual overview of the dataset and method is shown in Figure\ref{fig:framework}. 
Finally, to objectively evaluate the genre-matching degree between generated dances and given music, we propose a novel metric, named Genre-matching Score (GS).

Overall, our contributions can be summarized as follows:
\begin{itemize}
	\item We release FineDance, which is the largest 3D motion capture music-dance paired dataset with accurate full-body posture, containing 22 fine-grained genres. FineDance encourages the development of AI choreography, motion prior, and full-body reconstruction methods.%主语是数据集，而不是舞蹈The long duration, rich genre, and high-quality full-body dances
	% \vspace{-2.5mm}
	\item We present FineNet, which leverages expert networks and refine network to generate expressive full-body dances, and employs a cross-modal retrieval network to improve genre-matching scores.
	%可以！We propose FineNet, which combines the strengths of generative and synthesis methods,  effectively addresses the issue of monotonous and unnatural hand movements generated by previous methods.% overcoming the short duration problem of generative networks and the issues of genre mismatch and insufficient diversity of synthesis methods. %effectively addresses the issue of monotonous and unnatural hand movements generated by previous methods. 我觉得这句写在这里没有支撑，没有多大意义，所以我注释了。
	% \vspace{-2.5mm}
	\item Extensive quantitative experiments and user studies
	demonstrate that our approach can generate multiple different genre-matched dances from arbitrary music with natural and flexible hand movements.
	%Extensive quantitative experiments and user studies
	%demonstrate that our approach can generate multiple different genre-matched dances with nature and flexible hand movements.

\end{itemize}

\section{Related Works}
\label{sec:relatedworks}
\subsection{Choreography Dataset}
\label{subsec:Choreography_Dataset}
%We review the 3D choreography datasets. 
%Being lacking in more exact and accurate representations of human poses, 2D choreography datasets can only be used to generate ordinary dance videos, which are hard to adapt for the 3D dance generation task well, and limited in animation, entertainment, education and virtual reality applications.
Currently, the most popular 3D choreography dataset is AIST++ \cite{aist++}, which provides 5 hours of dance but does not have hand motions. AIST++ is  reconstructed through multi-view video. Therefore, there is inevitably a deviation between the generated 3D dance motions and the real motions because of the reconstruction error. Li \textit{et al.} \cite{danceformer} provided another major type of 3D dance dataset: modeling in software, which is obtained by experienced animators. However, this type of dataset lacks the authenticity of the dance. At present, the most accurate datasets are obtained from motion capture systems. GrooveNet dataset \cite{GrooveNet} only contains the electronic dance genre in 23 minutes sequence. Dance with Melody \cite{dancewithmelody} further constructed a 94-minute 3D dance dataset with 4 genres. %(waltz, tango, cha-cha, and rumba). 
Music2Dance \cite{Music2Dance} is a dataset with only one hour and 2 genres (modern and curtilage dance). Chen \textit{et al.} \cite{ChoreMaster} built a 9 hours dataset, but it only has four different dance genres. This genre partition does not match the discernment of professional dance artists. For example, Anime and K-pop do not count as specific dance genres.
In conclusion, these datasets have limitations in hand motion, duration, and the diversity of genres and dancers. Our dataset contains 14.6 hours with 22 genres, and is collected by 27 dancers.

\subsection{AI Choreography} 
\noindent\textbf{Synthesis based approach.} 
Early works\cite{ofli2011learn2dance, manfre2016automatic, berman2015kinetic, yang2020statistics, lee2013music} usually synthesis dances based on motion graphs and databases, which share one core idea: retrieving the most matching dance fragment for the given music clip, and splicing multiple fragments into a complete dance.
%implement machine choreography utilizing a synthesis method, which share one core idea:
In computer graphics, motion synthesis has long attracted a lot of attention as it can synthesize 3D actions from existing motion databases. Lamouret \textit{et al.} \cite{lamouret1996motion} first delivered this idea and proposed a prototype system to create new actions by cutting and pasting action fragments from an action database.
Arikan \textit{et al.} \cite{arikan2002interactive} formally introduced the concept of graph-based motion synthesis, transforming this problem into finding paths in a pre-built motion graph. Under such a framework, the task of synthesizing actions is usually regarded as finding the optimal path in a constructed motion graph. 
%剪篇幅
Similar utilizing this graph-based scheme, Kim \textit{et al.} \cite{kim2003rhythmic} made the first attempt to synthesize rhythmic movements by adding constraints connecting action beats and rhythmic patterns. 
Shiratori \textit{et al.} \cite{shiratori2006dancing} and Kim \textit{et al.} \cite{kim2006making} formally pointed the music-driven dance movement synthesis problem and further developed more complex rules to associate dance movement segments with input music segments.
%The advantage of this type of method is that its motion quality is high (from the database).
This type of method can synthesize dances that match the music style well. 
%However, since it is unable to learn the internal connections between music and dance, the synthesized dances is unable to match the rhythm well.
However, as it is unable to learn the internal connections between music and dance, the synthesized dances cannot match the rhythm well. In addition, because the dance fragments are all from the database, such methods have no ability to create new dance motions.
In this paper, we introduce the diversity generated technology into the synthesis method to maintain the advantages of synthesis and avoid the above problems.

\noindent\textbf{Generation based approach.} 
Generative Adversarial Network (GAN) \cite{GAN} and Variational Auto Encoder (VAE) \cite{VAE}  have been successfully applied to generate various data modalities, including image, motion, music, etc.  As a result, researchers also propose music-driven dance generation algorithms based on the general deep generation paradigm\cite{fragkiadaki2015recurrent, aist++, deepdance, aksan2019structured, danceformer, ghosh2017learning, villegas2018neural,he2023weaklysupervised,He2023Camouflaged}. 
Such methods can be divided into 2D and 3D solutions according to the dimension of the generated dance data. Lee \textit{et al.} \cite{dancingtomusic} proposed the first music-driven 2D dance generative network, which uses VAE to model dance units and GAN to loop to generate dance sequences. Since the human skeleton is a natural graph data, Ren \textit{et al.} \cite{Self_supervised_dance_synthesis} and Ferreira \textit{et al.} \cite{ferreira2021learning} adopted a Graph Convolutional Network to improve the spatial naturalness of the generated 2D dance movements. Both music and dance belong to sequence data. Therefore, Li \textit{et al.} \cite{aist++} used the Transformer \cite{Transformer} network with strong sequence modeling ability to design a music-driven 3D dance action generation network. Recently, diffusion-based networks succeed in text2motion generation\cite{tevet2022human,zhang2022motiondiffuse}.
Although these generative networks have the advantages, such as rhythm-matched and diversity, they neglect the quality of generated hand motions, and only generate motion fragments within several seconds.
Our method can generate expressive full-body dances due to our diffusion based expert nets, and use a retrieval based modal to enhance the genre-matching score and take the advantages of synthetic methods such as genre-matching, long-term stability.
%use the synthesis way to get long-term dance based on generated dance fragments to avoid the accumulation of errors.

%-------------------------------------------------------------------------
\section{FineDance Dataset}
\label{sec:MgM2D}

\begin{table*}[h!]
	\setlength\tabcolsep{3pt}
	\centering
	\resizebox{\textwidth}{!}{
		\begin{tabular}{lccccccccccc}
			\toprule [1pt] \noalign{\smallskip}
			Dataset &  Pos/Rot &  \makecell[c]{Joints\\num} & \makecell[c]{Hand\\joint}    & Genres & Mocap& \makecell[c]{RGB\\Views}   &Fbx&SMPL&Dancers&\makecell[c]{Total\\hours}& \makecell[c]{avg Sec\\per Seq}\\
			\hline\noalign{\smallskip}\noalign{\smallskip}
			GrooveNet\cite{GrooveNet} & \Checkmark/\XSolidBrush	  &	30	& \XSolidBrush&1&\Checkmark&1&\XSolidBrush&\XSolidBrush&1&0.38 & \textbf{690}\\
			Dance w/Melody\cite{dancewithmelody} &\Checkmark /\XSolidBrush  &21&\XSolidBrush&4&\Checkmark&4&\XSolidBrush&\XSolidBrush&-&1.6&92.5\\
			Music2Dance\cite{Music2Dance} & \Checkmark /\XSolidBrush  &55 & \Checkmark&2&\Checkmark&2&\XSolidBrush&\XSolidBrush&2&0.96&-\\
			EA-MUD\cite{deepdance} & \Checkmark /\Checkmark  &24 & \XSolidBrush&4&\XSolidBrush&4&\XSolidBrush&\XSolidBrush&-&0.35&73.8\\
			PhantomDance\cite{danceformer} & \Checkmark /\XSolidBrush  &24 & \XSolidBrush&13&\XSolidBrush&0&\XSolidBrush&\XSolidBrush&-&9.6&133.3\\
			AIST++\cite{aist++}  & \Checkmark /\Checkmark  &17/24 & \XSolidBrush&{10}&\XSolidBrush&\textbf{10}&\XSolidBrush&\Checkmark&-&5.2&13.3\\
			MMD\cite{ChoreMaster}  & \Checkmark /\Checkmark  &\textbf{52} & \Checkmark&4&\Checkmark&0&\Checkmark&\XSolidBrush&-&9.9&-\\
			\midrule [0.5pt]
			\textbf{FineDance (Ours)} &  \textbf{\Checkmark /\Checkmark}  &\textbf{52} & \textbf{\Checkmark}&\textbf{22} & \textbf{\Checkmark} & {2} & \textbf{\Checkmark}&\Checkmark&27 &\textbf{14.6}&152.3\\
			\noalign{\smallskip}
			\bottomrule [1pt] 
		\end{tabular}
	}   
	\vspace{0.1mm}
	\caption{Comparisons of 3D Dance Datasets. Pos and Ros means 3D position and Rotation information respectively.  Fbx (FilmBox) is one of the main 3D exchange formats as used by many 3D tools. "avg Sec per Seq" means the average seconds per sequence.}
	\label{tab:data_com}   
	\vspace{-1em}
\end{table*}

\begin{figure}[t]
	\centering
	\includegraphics[width=0.5\textwidth]{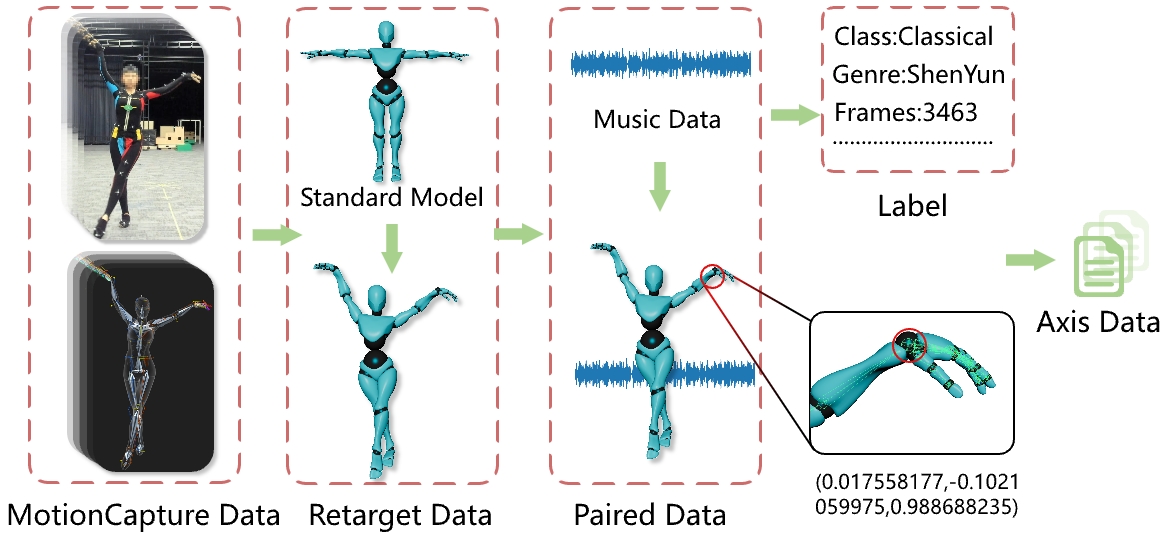}
	\caption{FineDance acquisition process. We capture the dancers' motions with the Vicon optical motion capture system. Then engineers retarget it to a standard skeleton in MotionBuilder. Dancers manually align the music with the dance motions and extract the skeletal information of the dance in Blender. }\vspace{-3mm}
	\label{fig:Datasets}
\end{figure}

The review of the choreography dataset is shown in Section \ref{subsec:Choreography_Dataset}, there are still few accessible large-scale motion capture choreography datasets even though many papers reported their choreography datasets. Meanwhile, most of these datasets are insufficient to train a diversified and generalized-well choreography model due to the limiting of dance genres, and poor music-dance pairings. Our FineDance can fill these gaps.% The lack of available finger motion data and the limited dance genres seriously hinder the development of dance generation. 这句是不是已经放在intro中了？不是原话，但是意思有了

\subsection{Data Acquisition and Analysis}
For the fine-grained dataset, we take the following regulations into account for our dataset acquisition. We give the flowchart of the process shown in Figure \ref{fig:Datasets}. We  summarized the comparison of FineDance and existing 3D dance datasets in Table \ref{tab:data_com}. 

% \begin{figure}[!bth]
	%     \centering
	%     \includegraphics[width=0.5\textwidth]{images/DataDistribute_super.png}
	%     \caption{Statistics of dance genres in FineDance. It contains 16 genres belonging to 4 classes.}
	%     \label{fig:dance_genre}
	% \end{figure}
\noindent\textbf{Fine-grained motions.} {So far, there are only 2 datasets containing finger motion, and the available data is less than 1.5 hours. Fine-grained motions are ignored in existing methods. For example, GrooveNet\cite{GrooveNet} only contains 30 body joints, and Dance with Melody\cite{dancewithmelody} has only 21 body joints, while EA-MUD\cite{deepdance}, PhantomDance\cite{danceformer} and AIST++\cite{aist++} all use 24 body joints of the SMPL\cite{SMPL} model. 
	
	Our data store the information of the skeleton joints in 3D space in each frame including fingers, which can help to improve the artistry and reality of the dance motion. For easy to utilize, we use the standard 52 joints to represent the dance data.
	
	\noindent\textbf{Fine-grained dance genres.} 
	Previous literature focuses on a few dance genres, such as GrooveNet\cite{GrooveNet}, Music2Dance\cite{Music2Dance}, MMD\cite{ChoreMaster} and Dance with Melody\cite{dancewithmelody}.  This dataset uses a rough genre classification strategy, which is a non-standardized dance division. 
	% For clarity, we show the dance genres of our dataset in Figure \ref{fig:dance_genre}. 
	
	We improve the diversity of our dataset from two aspects: more genres and more dancers. Our FineDance is reasonably classified under the advice of dance artists, covering hip-hop and Chinese classical dance more completely. To the best of our knowledge, it also includes folk dance motions for the first time, expanding the dance genres of the choreography dataset.  
	% As shown in Figure \ref{fig:dance_genre}, 
	Totally, FineDance has up to 22 genres of dance defined by professional dance artists. And we obtained more than 14 hours of data. It is worth noting that FineDance contains the most genres.
	Details are given in the supplementary materials. 
	
	% To make our dance data more in line with the criteria of dance artists, with the advice of professional dancers, we divided the dances according to genres as follows: Street Dance (Breaking, Locking, Hiphop, Popping, Urban and Jazz), Chinese Classical Dance (Hantang dynasties dance, Dunhuang Dance, Shenyun Dance and Kun Dance) and Folk Dance (Wei dance, Dai dance and Miao dance). Details in the supplementary materials. Meanwhile, professional dance artists also believe that Korean dance, China traditional Jazz and House dance do not belong to professional dance genres, and their dance motions mostly come from different dance styles, so we innovatively named them as mixed-style dances. To get realistic and accurate dance motions, we spent a lot of time on data repair and alignment, and finally we obtained more than 8 hours data.
	\noindent\textbf{Accurate posture.} 
	Currently, the largest available dataset is AIST++\cite{aist++}, which is collected by reconstructing 3D poses in multi-view videos. But the dance data is not real due to the reconstructing errors. Instead of reconstruction from the videos, ours is collected by a motion capture system, and  all dance motions and music are well paired. 
	
	In FineDance, all motions are captured by the Vicon optical motion capture system and retargeted to a standard skeleton in MotionBuilder by engineers. Therefore, FineDance can donate accurate postures.
	Moreover, FineDance will be the largest fully available 3D music-dance paired dataset, and it will be available.
	
	\noindent\textbf{Well-paired dance and music.} Dance fragments are strongly associated with the rhythm and style of music. However, due to the lack of enough well-paired data, the generative model is hard to fit the relevance of the motion rhythm and music rhythm. Therefore, we asked the professional dancers to pay attention to the matching of rhythm and style when dancing. 
	
	\noindent\textbf{Professional dancer.} 
	% To enhance the accuracy of dance performance, more professional dancers are necessary. 
	We invited 27 professional dancers, and each dancer was asked to dance to the music while his/her motions were captured utilizing the capture system.

	%\subsection{Dataset Analysis}
	%We analyze our dataset in many aspects, including genres, joints, duration, available/total seconds, etc. The comparisons of FineDance and existing 3D dance datasets are summarized in Table \ref{tab:data_com}. 

	%some of the datasets ignored the rotation information of bones and only used the position of bones, which is not conducive to the correct representation of dance motions. Also , this is certainly not in line with the realism of the dance.
	
	\begin{figure*}[t]
		\centering
		\includegraphics[width=\textwidth]{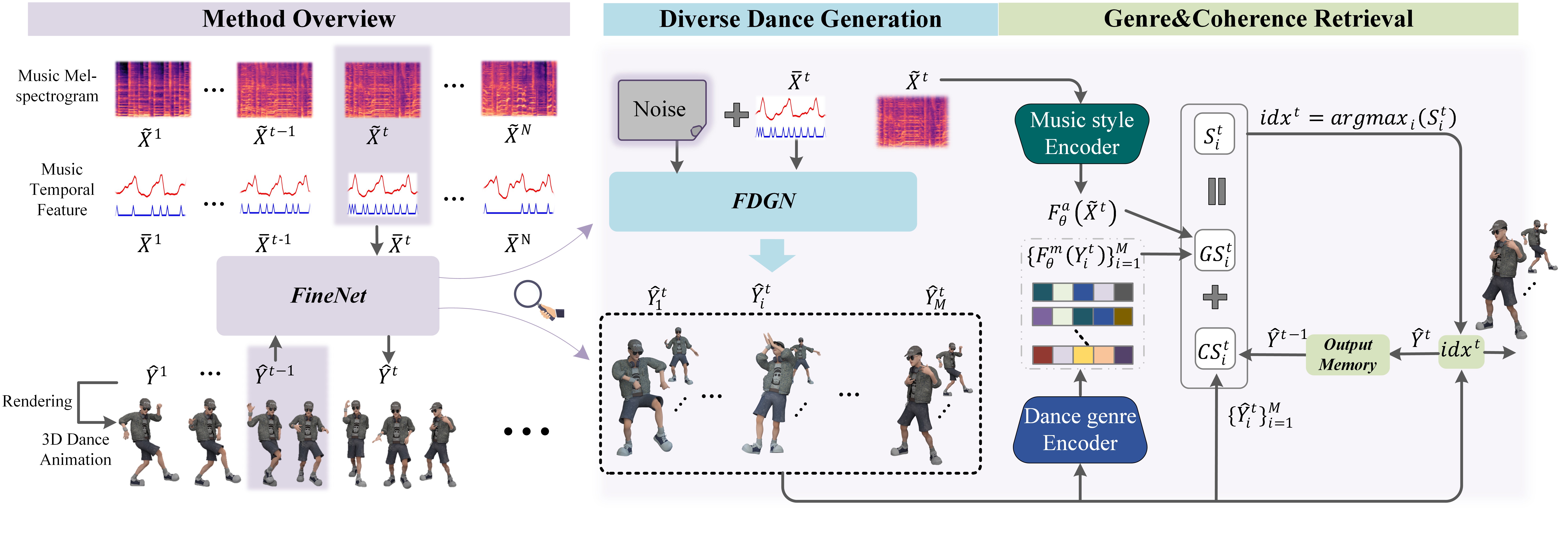}\vspace{-4mm}
		\caption{An overview of our framework. FineNet can iteratively generate and choreograph different dance fragments based on the mel-spectrogram and temporal features of music. FineNet consists of a diffusion-based Full body Dance Generation Network (FDGN) and a Genre$\&$Coherence retrieval module. The former is utilized to generate expressive and diverse full-body dance fragments; the latter is designed to retrieve the best matching dance fragments from the generated multiple dance fragments and synthesis them smoothly.}\vspace{-3mm}
		\label{fig:Method_Overview}
		% \vspace{-1.5em}
	\end{figure*}

	%-------------------------------------------------------------------------
	\section{FineNet}
	\label{sec:baseline}
	
	\subsection{Overall Framework}
	Given music of unknown style and arbitrary duration, our goal is to generate multiple different full-body dances. This task presents challenges in three areas: 
	(1) Full-body: Body and two hands motions in different spaces have different grains, using a single network to generate full-body motions like the previous method can lead to unreal and monotonous hand gestures. %检查的时候重点检查一下 冠词和单复数。这句两处应该加s,两处缺冠词。我已经加上了. 好！
	(2) Genre-matching: Making the generated dances consistent with the fine-grained genre is also challenging due to the modal gap between music and dance.
	(3) Long-term: Generating long-term novel motion is challenging because neural networks tend to accumulate errors over time. 
	%(1) Locally, the generated body and hand motions should be coordinated, diverse, and expressive; (2) Globally, the algorithm should be able to generate long-duration dances that are consistent with various musical genres.
	To address these issues, we propose FineNet, which comprises a Diffusion-based Full-body Diverse Dance Generation Network (FDGN) and a Genre $\&$ Coherent aware Retrieval Module (GCRM). The FDGN focuses on creating detailed dances with expressive movements, while the GCRM considers the overall choreography of the dance. FineNet cleverly combines generative and synthetic methods, making them complementary, much like the process of human choreography.
	%In addition, FineNet can generate multiple completely different dances by selecting different dance segments at the first time step. This feature provides the user with a range of creative options.
	%Moreover, by choosing different dance segments at the first time step, FineNet can generate multiple completely different dances.
	Furthermore, FineNet allows for the generation of multiple different dances by selecting distinct dance segments at the initial time step. This capability offers users a wide range of creative possibilities.
	
	The overall framework of our method is shown in Figure \ref{fig:Method_Overview}.
	First, the input music $X$ is split into 4-second clips $\{X^t\}_{t=1}^N$ without overlapping, and $N$ is the number of clips of the given music.%in the inference phase.这里简化后没有train phase了
	For each ${X}^{t}$, we use the Librosa toolbox \cite{librosa} to extract the temporal feature ${\bar{X}}^t\in\mathbb{R}^{T\times C^m}$, and the mel-spectrogram image  ${\widetilde{X}}^{t}\in\mathbb{R}^{W\times H\times3}$,
	where $T$ is the time length of a clip and $C^m$ is the channel dimension. $W$ and $H$ are the width and height of the image respectively, and 3 means the number of RGB channels.
	FineNet generates and retrieves the best dance fragment at each time step, and this process can be formulated as:
	\vspace{-1mm}
	\begin{equation}
		\begin{split}
			\begin{array}{lr}
				{\hat{Y}}^1 = FineNet({\bar{X}}^1, {\widetilde{X}}^{1}) &  \\
				{\hat{Y}}^t = FineNet({\bar{X}}^{t}, {\widetilde{X}}^{t}, \hat{Y}^{t-1}), 
			\end{array}
		\end{split}
	\end{equation}
	where $t\in \{2,3,\cdots,N\}$ is the current time step. ${\hat{Y}}^t\in\mathbb{R}^{V\times (T\times 3)}$ is the dance action fragment obtained by FineNet at time step t, and $V$ is the number of body and hand joints, ``3'' represents 3-dimensional axis angle and position of joints. 
	%---------------------------------------------------------------------
	%--------------------------------------------------------------------
	\subsection{Diverse Dance Generation}
	
	The previous generative models such as VAE\cite{siyao2022bailando} and GAN\cite{Self_supervised_dance_synthesis}, mostly directly link music embedding to dance embedding method, which results in the limited diversity of the generated dances. This is because high-level condensed features of music usually contain insufficient details to guide the network to generate different dances \cite{zhang2022motiondiffuse}. %\sout{{\color{blue}music} features typically contain insufficient details to guide the generation of subtly diﬀerent motion \cite{zhang2022motiondiffuse}.}
	
	To generate novel and diverse dances, we employ a diffusion-based model, FDGN. To make the network perceive the music rhythm  better, we feed the music temporal feature to FDGN ${\bar{X}}^{t}$ instead of ${X}^{t}$. Given ${\bar{X}}^{t}$ and different noise sampled from $\mathcal{N}(0,I)$,  FDGN can generate $M$ distinctive dance fragments, represented as ${{\{{\hat{Y}}_i^t}\}}_{i=1}^M$.

	\subsection{FDGN: Full-body Dance Generation Network}
	\begin{figure}[t]
		\centering
		\includegraphics[width=0.48\textwidth]{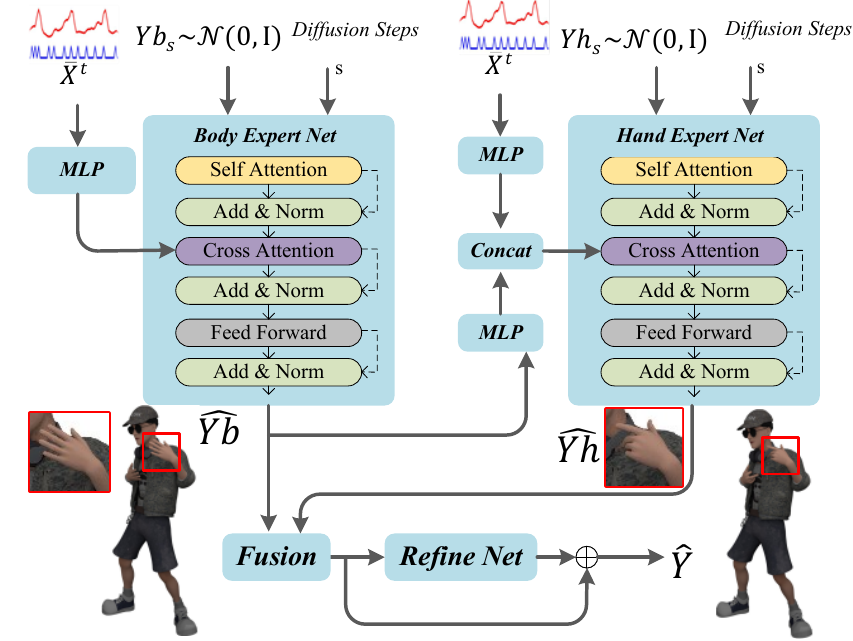}\vspace{-2mm}
		\caption{Full-body Dance Generation Network."$\bigoplus$" means add.}\vspace{-3mm}
		\label{fig:FDGN}
	\end{figure}
	We propose FDGN, a network for generating full body dance motions. Compared to previous models, FDGN can produce more realistic, natural, and expressive hand motions that coordinate well with body motion and music styles. As shown in Figure \ref{fig:FDGN}, we use two expert networks to generate body and hand motion separately. This is based on two observations: (1) The range of motion of the body and hand is obviously different and belongs to different feature spaces. Therefore, using a single network to generate full-body dances can result in unnatural hand motions. (2) In dance generation, music rhythm mainly coordinates with the limbs, and hand motion should be consistent with the body and music style.
	%In order to make the hand movements more coordinated with the body motion and music style. We propose FDGN, which is consist of diffusion based body/hand expert net and a refine net.
	%In the field of full-body motion capture, most researchers usually use expert networks to capture body and hand motion separately. Inspired by this, we also designed body expert net and hand expert net based on Diffusion, which are used to generate body and hand motions respectively.
	% The body and hand expert is based on human diffusion model (MDM)\cite{tevet2022human}. They 
	The structure of body/hand expert net is modified by MDM\cite{tevet2022human} and EDGE\cite{tseng2023edge} . The training process of body expert net can be formulated as:
	% 篇幅
	\vspace{-3.5mm}
	\begin{equation}
		\mathcal{L}_d=\mathrm{E}_{s \in[1, S], Yb_0 \sim q\left(Yb_0\right)}[\left\|{Yb_0}-{BEN}\left(Yb_s, s, \bar{X}\right)\right\|_2^2],
	\end{equation}
	%\end{footnotesize} 
	where $Yb_0$ is the label dance data, $BEN(~)$ means Body Expert Net, $s$ is the number of diffusion steps. %篇幅($S=1000$ for our experiments).
	The refine net is consisted of a spatial convolution network and a refine gate unit, which is used to assemble two parts naturally.

	\subsection{Genre$\&$Coherent aware Retrieval Module}
	\label{subsec:GCRM}
	Thanks to FDGN's diverse generative capabilities, we can generate $M$ candidate dance fragments $\{\hat{Y}^t_i\}_{i=1}^M$ for the $t-th$ music clip ${\{\bar{X}^t\}}$.
	% Given a piece of music, we split it into multiple music clips $\{{X^t}\}_{t=1}^{N^{\prime}}$ per 4 seconds without overlap. For each music {\color{blue}clip}  $X^t$, we generate $M$ candidate dance fragments $\{\hat{Y}^t_i\}_{i=1}^M$ by D3Net. 
	To search the genre-matched candidate dance fragment and synthesis them coherently, we propose Genre$\&$Coherent aware Retrieval Module (GCRM), which calculates the Genre Matching Score (GS) of the current music clip and the candidate dance fragments, and the coherent score of the previous selected dance fragment and the current candidate dance fragments. Finally, we choreograph the complete dance by combining the two scores.
	
	% Based on the diversity dance generation capability of D4Net, we can obtain $M$ candidate dance clips $\{\hat{Y}^t_i\}_{i=1}^M$ form each music clip $X^t$.

	%To achieve this goal, we propose a scoring strategy which contains the stylized dance matching score and the coherent score.
	
	\noindent\textbf{Genre matching score.} We use GS to evaluate the matching degree of the style of music clip  $\widetilde{X^t}$ and the genres of generated candidate dance fragments $\{\hat{Y}^t_i\}_{i=1}^M$. However, music and dance are different data modalities, and it is difficult to calculate the genre similarity directly. To cover this problem, we propose a cross-modal retrieval network. 
	
	We utilize two models (one music style encoder $\mathcal{F}_\theta^m(\cdot)$ and one dance genre encoder $\mathcal{F}_\theta^d(\cdot)$) to encode different modalities into one embedding space. The music style encoder is the backbone of AST \cite{gong2021ast} and the dance genre encoder is the backbone of AGCN \cite{shi2019two}. 
	We extract the mel-spectrogram features $\widetilde{X^t}$ of music $X^t$ and send it to the music encoder. 
	For each $\{\widetilde{X^t}\}_{t=1}^{N}$, we compute its genre-matching score with candidate dance fragments generated by FDGN. %The genre-matching score is implemented with a cosine similarity function, which can be expressed by the following formula
	We utilize the cosine similarity as the genre matching score. 
	So the genre matching score between $\widetilde{X^t}$ and $\{\hat{Y}^t_i\}_{i=1}^M$ can be formulated as:
	%\vspace{-3mm}
	% we  cosine similarity as the matching score to measure the matching of music and dance. The similarity calculation formula is as follows:
	\begin{equation}
		{GS}_{i}^t = s(\widetilde{X^t}, \hat{Y}^t_i) = \frac{\mathcal{F}_\theta^m(\widetilde{X^t})^T \mathcal{F}_\theta^d(Y_i^t)}{||\mathcal{F}_\theta^m(\widetilde{X^t})||\times ||\mathcal{F}_\theta^d(Y_i^t)||},
		\label{formula:GS}
	\end{equation}
	% \vspace{-3mm}
	%The training setup is shown in Figure \ref{fig:match_train}. 
	We train the above two models by a cosine loss $\mathcal{L}_{cos}$ as a retrieval task, which can be formulated as: 
	%\vspace{-2mm}
	%The difficulty in evaluating genre-matching score between dance and music is the domain gap of different data modals. To address this issue, we propose a cross modal retrieval network to calculate it.
	%The genre-matching of generated dance with the given music is a crucial aspect of choreography evaluation. However, previous research did not experiment with appropriate metrics for this purpose due to the challenge of comparing two different modalities.
	%---------------好句子-------------
	%这一段注释了，篇幅变小了点， 过会儿我挪到方法部分
	
	%\sout{We generate multiple dance fragments for a music piece in the inference phase of the generation network and aim to find the one which is the most related to the music style in those generated dance fragments. }
	% Music and dance are different data modalities, and it is difficult to calculate the genre similarity directly. To cover with this problem, we propose a new metric named Genre Matching Score (GS), with a neural network to calculate it.
	
	% \begin{equation}
		%   \begin{split}
			%   &\mathcal{L}_{cos} = \left\{
			%             \begin{array}{lr}
				%              1-s(\widetilde{X_i}, {Y_j}),if\ y=1 &  \\
				%              max(0,s(\widetilde{X_i}, {Y_j})-\eta), if\ y=-1
				%              \end{array}
			% \right.,   
			%   \end{split}
		%   \end{equation}
	\begin{equation}
		\begin{split}
			&\mathcal{L}_{cos} = 
			y(1-s(\widetilde{X_i}, {Y_j}))+(1-y)(max(0,s(\widetilde{X_i}, {Y_j})),
		\end{split}
	\end{equation}
	where $\widetilde{X_i}$ is the MFCCs feature of music clip, ${Y_j}$ is a dance fragment. $y=1$ when  $\widetilde{X_i}$ and ${Y_j}$  are genre-matched, otherwise $y=0$.
	
	% debug!!!图4省略！
	% \begin{figure}[!bth]
		%     \centering
		%     \includegraphics[width=0.5\textwidth]{images/matchtrain.pdf}
		%     \caption{The training setup of genre matching score calculation network.}
		%     \label{fig:match_train}
		% \end{figure}

	\noindent\textbf{Coherent score.}
	%Our synthesis strategy is not only the matching of style, but also 
	%\sout{We utilize the feature similarity of the tail and head of the two temporal fragments to measure the coherences. }
	We take the L2 distance of the start and end states of the two dance segments as the coherence score. To make the transition smoother, we cut 5 frames of the last temporal clip ($t-1$) best matching fragment $\hat{Y_b}^{t-1}$ on the tail and cut the start 5 frames of the $M$ candidate dance fragments $\{\hat{Y_i^{t}}\}_{i=1}^M$ at the current time step $t$.
	These cut frames are finally filled with a linear interpolation algorithm. 
	\vspace{-2mm}
	\begin{equation}
		\begin{split}
			\begin{array}{lr}
				{CS}_{i}^1 =0  \\
				{CS}_{i}^t ={-|| Y_b^{t-1}[-5,:] - Y_i^{t}[5,:] ||_2},
			\end{array}
		\end{split}
	\end{equation}
	where $t\in \{2,3,\cdots,N\}$ is the time step.
	
	Combining GS and CS, GCRM can find the best ${\hat{Y}}^t$ from the candidate dance segments $\{Y_i^t\}_{i=1}^M$ as the output at time step $t$:
	\vspace{-3mm}
	\begin{equation}
		\begin{split}
			\begin{array}{lr}
				{idx}^t={\rm argmax}_{i}{\left(\alpha{GS}_i^t+\beta{CS}_i^t\right)},  \\
				{\hat{Y}}^t:={\hat{Y}}_{{idx}^t}^t, {\hat{Y}}_{{idx}^t}^t\in \{Y_i^t\}_{i=1}^M,
			\end{array}
		\end{split}
	\end{equation}
	where $\alpha$ and $\beta$ are weight parameters.
	For complete music $X$, FineNet outputs the dance fragments step by step, and the final result is: $\hat{Y}=[\hat{Y}^1,\hat{Y}^2,...,\hat{Y}^T]$. 
	Furthermore, by choosing different $\hat{Y}^1$ at step 1, FineNet can generate multiple dances with excellent diversity.
	
	%\sout{\textbf{Long-sequence dance stitching}
		%For a piece of music, we have obtained the index $\{idx^t\}_{t=1}^T$ of each split according to the style and coherent of the whole music. The dance fragment of music piece $t$ can be gathered: $\hat{Y}^t=\mathcal{G}(\{Y_i^{t}\}_{i=1}^M,idx^t)$, where $\mathcal{G}(Set,idx)$ denotes the function to extract the corresponding element of the $Set$ according to the index $idx$. We directly spliced all the dance fragments in chronological order: $\hat{Y}=[\hat{Y}^1,\hat{Y}^2,...,\hat{Y}^T]$.}

	%-------------------------------------
	
	\section{Experiments}
	\label{sec:exper}
	%这两句重复吧
	
	\subsection{Experimental Setup}
	\noindent\textbf{Data Preprocessing.} 
	We spilt FineDance dataset into train, val and test sets in two ways: FineDance@Genre and  FineDance@Dancer. The test set of FineDance@Genre includes a broader range of dance genres, but the same dancer appear in  train/val/test set. FineDance@Dancer means the train/val/test set divided by different dancers, which test set contains fewer dance genres, yet the same dancer won't appear in different sets. We only reported the results of test set on FineDance@Genre in this paper, the details of dataset split can be found in our supplementary materials.
	%We spilt FineDance dataset in two ways, cross dancer and cross genre.
	%We split our dataset into train, val and test set, and report the performance on the test set only.
	% 
	Each music and paired dance are only present in one set. %The training set contains 22 genres and 50708 seconds of dance-music paired data. The test set contains 20 genres, and 1993 seconds of dance-music paired data. 
	%We split the dataset to make sure that the test set contains most of the genres and different duration from 43 seconds to 193 seconds of dance data. 
	%$\rm{D^3GNET}$
	\begin{figure}[t]
		\centering
		\includegraphics[width=0.48\textwidth]{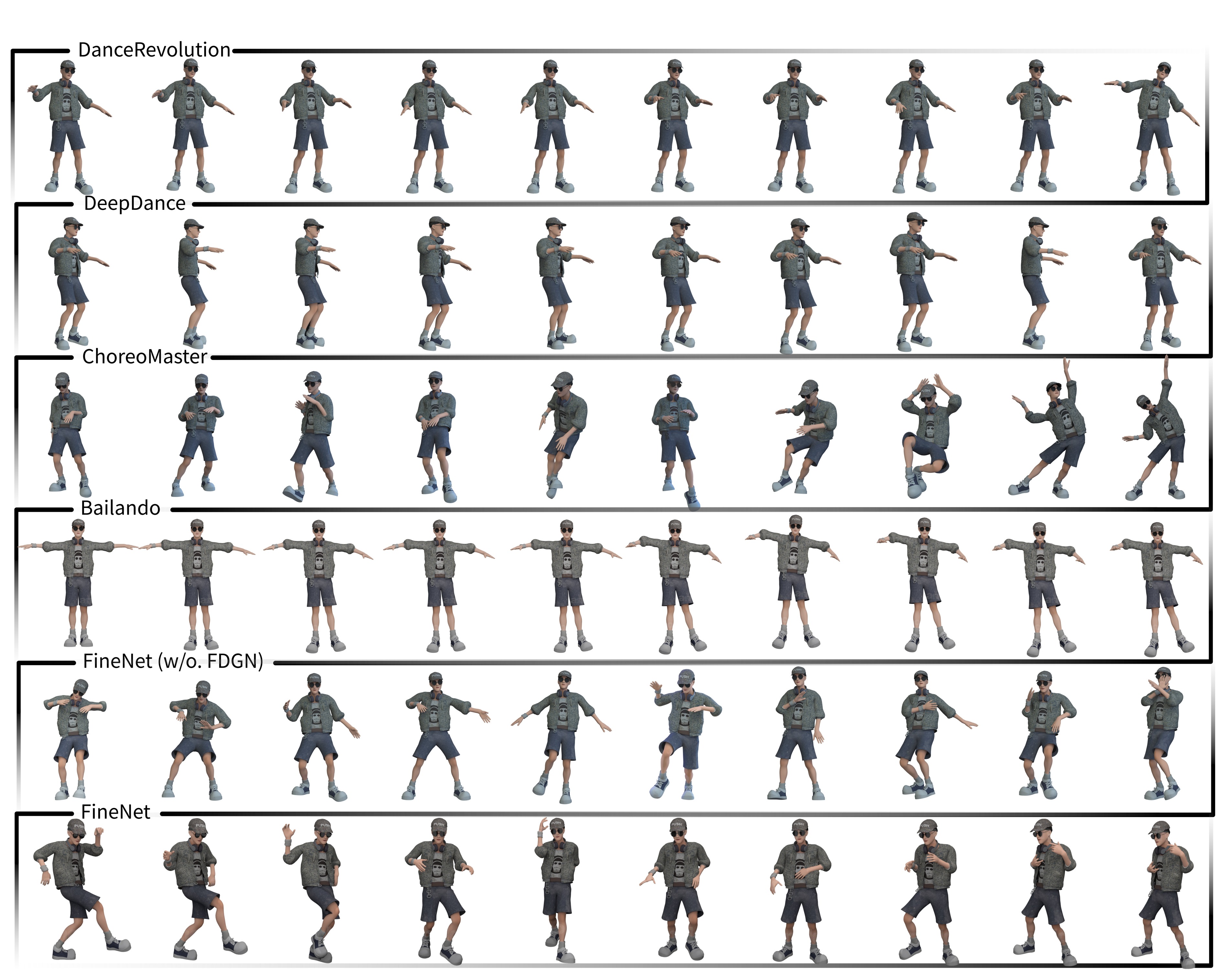}%\vspace{-2mm}
		\caption{Qualitative result comparisons for a Jazz song.}\vspace{-3mm}
		\label{fig:qualitative}
		\vspace{-1em}
	\end{figure}
	For all the dance fragments, we combine the 3-dim axis angle vector representation for all 52 joints, along with a 3-dim global position vector, resulting in 159-dim motion features. For all the music clips, we use Librosa \cite{librosa} to extract the 35-dim music temporal features. 
	%the following features: 1-dim envelope, 20-dim MFCC, 12-dim chroma, 1-dim one-hot peaks and 1-dim one-hot beats, resulting in a
	We also extract the mel-spectrogram of the music clips with Librosa and resized it to $224\times 224\times 3$. During extracting audio features, the sampling rate is 76,800Hz and hop size is 512. %Then we have 30 frames of feature per second for both motion and music sequences.%Finally, following DeepDance \cite{deepdance}, we stack the feature of 5 pieces for one music frame which results in a 175-dim vector. 
	
	% We conduct sufficient experiments on 
	% test set to verify the effectiveness of our method.
	\begin{table*}
		\resizebox{\linewidth}{!}{
			\begin{tabular}{lcccccccccccc}
				\toprule [1.5pt]
				\noalign{\smallskip}
				\multirow{2}{*}{Method}& \multicolumn{6}{c}{FineDance Dataset} & \multicolumn{6}{c}{AIST++ Dataset} \\  \cmidrule(lr){2-7} \cmidrule(lr){8-13}
				\noalign{\smallskip}
				& $FID\downarrow$ &  $FID_h\downarrow$ &$Div\uparrow$ & $Div_h\uparrow$ &$MM\uparrow$&  $GS\uparrow$ & $FID\downarrow$ & $FID_h\downarrow$& $Div\uparrow$ & $Div_h\uparrow$&  $MM\uparrow$&    $GS\uparrow$\\
				\noalign{\smallskip}\midrule
				\noalign{\smallskip}
				Ground Truth&   /	& /& 6.28  &	3.48&/	& 0.71 & / & /&9.07&/ &/ & 0.77\\\midrule
				ChoreoMaster\# & 1.92& 0.61 & \textbf{6.18}  &3.33 &	12.20	& 0.39 & 2.21 &/& 9.38&/&30.71&0.71\\
				DanceRevolution* &7.44 &3.21 &3.92 & 1.80&4.22&0.30 & 6.05&/&7.67&/&12.88&0.75 \\
				DeepDance* &  5.77 & 1.95&5.07 & 3.50&0.73 & 0.36& 25.78&/&8.98&/&3.08&0.73\\
				Bailando* &  4.77 & 2.24 &3.09 & 1.12 &2.33 & 0.29& 17.45&/&9.44&/&2.58&0.74\\
				\midrule [0.5pt]
				\textbf{FineNet (w/o. FDGN)} &  1.90 & 1.20&5.87 &3.40 &12.86 & 0.62& 2.11&/ &9.13&/ &31.09 & 0.79\\
				\textbf{FineNet } &  \textbf{1.66} & \textbf{0.48}&5.99 &\textbf{3.59} &\textbf{16.72} & \textbf{0.74}& \textbf{2.05}&/ &\textbf{9.94}&/ &\textbf{33.67} & \textbf{0.80}\\
				\bottomrule [1.5pt] 
				\noalign{\smallskip}
		\end{tabular}}
		\caption{Compared with SOTAs. \# means we reproduce the code, * means we use the published code.}
		\label{tab:results} 
		\vspace{-1em}
	\end{table*}
	
	% \begin{table}[th]
		% %\resizebox{0.5\textwidth}{!}{
			% 	\begin{tabular}{lcccc}
				% 		\toprule [1pt] 
				% 		Method & FID$\downarrow$ &   Div$\uparrow$ &  MM$\uparrow$&  GS$\uparrow$\\
				% 		\noalign{\smallskip}\midrule [0.5pt]
				% 		ChoreoMaster &   2.85	& 6.22  &	12.02	& 0.27\\
				% 		DanceRevolution  &  1.68 &\textbf{6.34} &11.51 & 0.46\\
				%   		Bailando  &  1.68 &\textbf{6.34} &11.51 & 0.46\\
				% 		\textbf{FineNet} &\textbf{1.66} &5.99 &\textbf{16.72} & \textbf{0.74}\\
				% 		\bottomrule [1pt]
				%                \noalign{\smallskip}
				% 	\end{tabular}
			% 	\caption{Compared with SOTAs. Tested on FineDance@Dancer dataset.}
			% 	\label{tab:Result_DancerDataset}  
			%         \vspace{-1em}
			% \end{table}
		
		\noindent\textbf{Implementation Details.}  In FDGN, we build 3 MLP layers to encode the music and body features. We use the transformer layer as the backbone of the body/hand expert net. The refine net consists of a 1-D convolution layer and a learnable weight parameter. The total epoch, learning rate, and batch size are set as $200$, 2$e^{-4}$, $2048$. In the GCRM, the $\alpha$, $\beta$ are set as $1.0$ and $0.5$ respectively. %All networks in our experiment are trained on  8 Nvidia A100 GPU.
		
		\noindent\textbf{Evaluation Metrics.}   (1) \textbf{FID score}. Fréchet inception distance (FID) is widely used to measure how close the distribution of the generated dances is to that of the ground truth. Similar to Lee \textit{et al.}\cite{dancingtomusic}, we trained a style classifier in our dataset to extract motion features, and then use the features to calculate FID.  (2) \textbf{Diversity}. We follow Lee \textit{et al.} \cite{dancingtomusic} to evaluate the average feature distance between generated dances for different input music. The same feature extractor used in FID is used again. (3)\textbf{Hand FID score} and \textbf{Hand Diversity}. Similarly, we extract hand motion features and calculate the FID and diversity for hand motion. (4) \textbf{Multimodality}. We follow Lee \textit{et al.}\cite{dancingtomusic} to evaluate the average feature distance between the 10 choreography versions of every music. This metric measures the model's ability to generate different dances for the same music. (5) \textbf{Genre Matching Score.} We evaluate the average genre matching score between generated dance and the input music using the genre matching score calculation network mentioned in section \ref{subsec:GCRM}. The genre matching score is defined as Eq. (\ref{formula:GS}).
		%\begin{equation}
		%    \mathcal{GS}=\frac{1}{N}\sum_{j=1}^{N}||M_j-D_j||_2^2~,
		%\end{equation}
		%where $N$ means the number of the song clips, $M_j$ and $D_j$ means the embedding vector of the $j-th$ music and dance clip.
		
		\subsection{Quantitative and Qualitative Evaluation}
		
		\noindent\textbf{Compared methods.} We compare our method with several generation-based methods and one synthesis-based method.  ChoreoMaster \cite{ChoreMaster} is a dance synthesis method with style embedding network and graph-based motion synthesis. Since the code is not available, we reproduced the code according to the paper. DanceRevolution \cite{dancerevolution} is a generation method using Tansformer to generate long-term dances. DeepDance \cite{GAN} is a generation method with a GAN-based cross-modal association framework. Bailando \cite{siyao2022bailando} is a generation method with VAE and an actor-critic generative pre-trained transformer model. All the methods are tested on FineDance and AIST++ to evaluate the comprehensive choreography ability of FineNet.
		
		\noindent\textbf{Results and analysis.} For all methods, we generate 10 versions of the dance for each song in the test set, and take the paired real dance as the Ground Truth. As shown in Table \ref{tab:results}, our method gets the best performance in all evaluation metrics except Diversity on both datasets. ChoreoMaster gets better on Diversity because it must retrieve all the fragments in the whole training set but no new motion is created. Specifically, on Multimodality (MM), FineNet gains 4.52 improvements compared to ChoreoMaster. On GS, FineNet is 0.03 higher than the ground truth. These two metrics show FineNet can generate diverse and genre-matched dance. Besides, FineNet gets the highest Hand FID and Hand Diversity which shows our method can generate real and diverse hand motion. %FineNet also increase 0.26 against ChoreoMaster on FID which means the dances generated by FDGN are more close to the RealDance.
		%These results verify our motivation of FDGN on diversity and GCRM method on genre matching. 
		
		Qualitative results are shown in Figure \ref{fig:qualitative}. DanceRevelution, DeepDance and Bailando generated almost identical stiff motions (refer to row 1 and row 2). ChoreoMaster generates classical dance motions in the last two images that do not match the Jazz music. But the generated dance motions of FineNet are diverse and in line with the Jazz style. The hand motion generated by FineNet W/ FDGN is more real. The corresponding videos refer to supplementary material.
		%
		%For FID, our method gets the highest score which shows FDGN can generate more real dance than existing generative networks.
		
		\subsection{Ablation Study}
		We replace different components of our method, and conduct experiments to demonstrate the effectiveness of each component on the test set of FineDance.
		
		\noindent\textbf{Infulence of the FDGN.} In order to verify whether FDGN can make hand movements more natural and flexible. We modified FDGN as a single network that generates body and hand motion simultaneously. We then employed $FID_h$ and $DIV_h$ metrics to evaluate the realism and diversity of the resulting hand motion.  As shown in Table\ref{tab:results}, all the quantitative metrics for FineNet w/o. FDGN becomes worse on both datasets. Specifically, on Hand Fid and Hand Diversity, FineNet w/o. FDGN drops by 0.72 and 0.14. The huge increase in $FID_h$ is mainly because FDGN generates hand movements with complete body information, making the hand motions more coordinated with the body.
		
		%\noindent\textbf{Infulence of the FDGN.} We modified FDGN as a single net to generate body and hand motion at the same time, named FineNet w/o expert net. As shown in Table\ref{tab:results}, all the quantitative metrics for varient w/o. FDGN become worse on both datasets. Specifically, on Hand Fid and Hand Diversity, varient w/o. FDGN drop by 0.72 and 0.14, which shows the effectiveness of FDGN for hand motion generation.
		
		\begin{table}[th]
			%\resizebox{0.5\textwidth}{!}{
				\begin{tabular}{lcccccccc}
					\toprule [1pt] 
					Strategy &   FID $\downarrow$ &   Diversity $\uparrow$ &  MM $\uparrow$ &    GS $\uparrow$\\
					\noalign{\smallskip}\midrule [0.5pt]
					\noalign{\smallskip}
					Ground Truth&   /	& 6.28  &	/	& 0.71\\
					\midrule [0.5pt]
					FDGN-G &   2.85	& 6.22  &	12.02	& 0.27\\
					FDGN-C  &  1.68 &\textbf{6.34} &11.51 & 0.46\\
					\textbf{FineNet} &\textbf{1.66} &5.99 &\textbf{16.72} & \textbf{0.74}\\
					\bottomrule [1pt]
					\noalign{\smallskip}
				\end{tabular}
				%}
			\caption{Ablation study on different choreography strategies.}
			\label{tab:ablation}  
			%\vspace{-1em}
		\end{table}
		\noindent\textbf{Choreography Strategy.} To verify the proposed generative-synthesis strategy, we use the following strategies to generate 3D dances: directly generating long-term 3D dances by FDGN (FDGN-G); generating 4s dance fragments according to the music clip by FDGN and concatenating these fragments together (FDGN-C), as Table \ref{tab:ablation} shows.
		
		FineNet performs better than others in FID, MM, and GS. For FDGN-G, a long music clip will cause the FID and GS to drop dramatically. %FDGN-C gets a higher Diversity score than the RealDance and FineNet, 
		FineNet achieves a lower Diversity score than FDGN-C because while retrieving the GCRM gives up some dances whose genre doesn't match. For MM and GS, FineNet increases by 5.21 and 0.28 compared to FDGN-C, which demonstrates that GCRM can increase the ability to generate diverse and genre-matched dance for the same song.
		
		\begin{table}[th]
			\resizebox{0.5\textwidth}{!}{
				\begin{tabular}{lcccccccc}
					\toprule [1pt] 
					Generation Model &   FID $\downarrow$ &  Diversity $\uparrow$ &  MM $\uparrow$ &    GS $\uparrow$\\
					\noalign{\smallskip}\hline\noalign{\smallskip}
					DeepDance &4.91 &4.47 &1.15&0.32\\
					DanceRevolution &6.99 &3.51 &6.64&0.37\\
					\textbf{FDGN} &\textbf{1.66} &\textbf{5.99} &\textbf{16.72} & \textbf{0.74}\\
					\bottomrule [1pt] 
					\noalign{\smallskip}
				\end{tabular}
			}
			\caption{Ablation study on different generation networks.}
			\label{tab:2}   
			\vspace{-3mm}
		\end{table}
		\noindent\textbf{Influence of the generation model.} We choose DanceRevolution, DeepDance as the backbone of the generative model, and get the whole 3D dances through our GCRM.
		%. We use these networks to generate 10 dance fragments for all clips of every song in the test set, and get a whole 3D dance through our GCRM.  
		All the results are shown in Table \ref{tab:2}. Our FDGN performs best, which shows excellent dance generation ability. Besides, compared the results of Table \ref{tab:results} and Table \ref{tab:2},  where DeepDance and DanceRevolution are inserted into our method in Table \ref{tab:2}, and their performances can be improved on MM, FID and GS. These results show our GRCM is scalable.
		
		% \begin{table*}[ht]
			% 	\begin{tabular}{lcccccccc}
				% 		\toprule [1pt] 
				% 		Method &  Style matching &  Rhythm matching &    Diversity & Comprehensive artistry & Average\\
				% 		\noalign{\smallskip}\hline\noalign{\smallskip}
				% 		RealDance(ground truth) &   94.1	& 96.3  &	95.5	& 97.9&96.0\\
				% 		\midrule [0.5pt]
				% 		ChoreoMaster &   64.2	& 59.7  &	77.2	& 69.8&67.7\\
				% 		DanceRevolution &33.2 &30.7 &47.4&31.4&36.7\\
				% 		DeepDance &  17.2 &22.1 &24.4 & 25.6&22.3\\
				% 				\midrule [0.5pt]
				% 		\textbf{FineNet(Ours)} &  \textbf{79.9} &\textbf{70.3} &\textbf{81.2} & \textbf{76.4}&\textbf{77.0} \\
				% 		\bottomrule [1pt] 
				% 	\end{tabular}
			% 	\caption{result of the user study.}
			% 	\label{tab:user_study}   
			% \end{table*}
		
		\subsection{User Study}  %As a traditional art form, the subjective feelings to the generation dances are also important. 
		We invite 30 participants, including 15 dancers. Everyone watches 5 dances with a duration of 32 seconds. They are asked to evaluate the dances in  2 aspects: Effective duration and Artistry.  
		\begin{table}[t]
			\centering
			%\resizebox{0.5\textwidth}{!}{
				\begin{tabular}{lcccccccc}
					\toprule [1pt] 
					Method &   Duration $\uparrow$& Type\\
					\noalign{\smallskip}\hline\noalign{\smallskip}
					Ground Truth & 30.7s&/\\
					\midrule [0.5pt]
					ChoreoMaster &  26.9s&synthesis\\
					DanceRevolution &5.4s&generation\\
					DeepDance &  2.2s&generation\\
					Baildando & 3.1s&generation\\
					\midrule [0.5pt]
					%FDGN(Ours) & 8.7s\\
					\textbf{FineNet (Ours)} & \textbf{28.2s}&generation\&synthesis \\
					\bottomrule [1pt] 
					\noalign{\smallskip}
				\end{tabular}
				%}
			\caption{User study on effective duration.}
			\label{tab:duration} 
			% \vspace{-1em}
		\end{table}
		\begin{table}[t]
			\centering
			\begin{tabular}{lcccccccc}
				\toprule [1pt]
				Method &  GM &  RM &    Div. & CA & Avg\\
				\noalign{\smallskip}\hline\noalign{\smallskip}
				Ground Truth &   94.1	& 96.3  &	95.5	& 97.9&96.0\\
				\midrule [0.5pt]
				ChoreoMaster &   64.2	& 59.7  &	77.2	& 69.8&67.7\\
				DanceRevolution &33.2 &30.7 &47.4&31.4&36.7\\
				DeepDance &  17.2 &22.1 &24.4 & 25.6&22.3\\
				Bailando &  23.2 &33.5&39.4 & 21.6&30.1\\
				\midrule [0.5pt]
				\textbf{FineNet (Ours)} &  \textbf{79.9} &\textbf{70.3} &\textbf{81.2} & \textbf{76.4}&\textbf{77.0} \\
				\bottomrule [1pt] 
				\noalign{\smallskip}
			\end{tabular}
			\caption{Result of the user study. Avg denotes the average score.}
			\label{tab:user_study} 
			\vspace{-3mm}
		\end{table}
		
		\noindent\textbf{Effective duration.} Each participant has to judge the time interval from the beginning until abnormal actions occur, such as abnormal shaking, motion freezing, or excessive joint distortion.
		As Table \ref{tab:duration} shows, synthesis methods generally performs better than generation methods, generation methods can only generate effective dances for less than 10 seconds. These results show that generation methods are limited in effective duration. But, ours is a generative-synthesis method, and achieves the best performance. 
		
		\noindent\textbf{Artistry.} We measure the artistry of the dances from the following dimensions: \textit{genre matching (GM), rhythm matching (RM), diversity (Div.), and comprehensive artistry (CA)}. The full score of every dimension is 100. As shown in Table \ref{tab:user_study}, synthesis methods generally perform better than generation methods, and our method gets a 77.0 average score which clearly surpasses the baselines. However, compared to the ground truth, all the methods still need to improve especially in rhythm matching and Comprehensive artistry. 
		
		%From user studies, the ground truth dances are widely praised by testers and achieve a average score of 96, indicating the high quality of the FineDance.
		%Both effective duration and artistry of ground truth are widely praised by testers,  which demonstrate the dance quality of our FineDance.
		
		\section{Conclusions}
		% In this paper, we propose a synthesis-optimized AI choreography algorithm, which combines the advantages of the generative network and the synthesis algorithm, so that the generated dance can be long enough while maintaining the beauty and rhythm of details. At the same time, we also conduct a large number of experiments to prove the effectiveness of the method.
		In this paper, we propose a large-scale, high-quality 3D dance dataset (FineDance) for music-driven dance generation, which records professional and abundant dance genres with accurate and fine-grained hand motions.
		Meanwhile, we also propose a choreography Network (FineNet), which can generate multiple diverse genre-matched dances with flexible hand movements.
		%which combines the advantages of generative networks and synthetic algorithms can generate fine-grained and long-term dances, and we solve the problem of monotonous hand motion. 
		Furthermore, we propose a new metric to evaluate the genre matching degree between music and dance.  Quantitative and qualitative results show that FineNet can generate long-term, diverse, and genre-matched dances from given music.
		
		\section*{Acknowledgment}
		We would like to express our sincere gratitude to Yan Zhang \href{https://yz-cnsdqz.github.io/}{(https://yz-cnsdqz.github.io/)} and Yulun Zhang \href{https://yulunzhang.com/}{(https://yulunzhang.com/)} for their invaluable guidance and insights during the course of our research.  %Their expertise and advices were instrumental in shaping our study. 
		
		This work was supported in part by the Shenzhen Key Laboratory of next generation interactive media innovative technology (No.ZDSYS20210623092001004), in part by the China Postdoctoral Science Foundation (No.2023M731957), in part by the National Natural Science Foundation of China under Grant 62206153, Young Elite Scientists Sponsorship Program by CAST (No. 2023QNRC002)
		
		{\small
			\bibliographystyle{ieee_fullname}
			\normalem
			\bibliography{egbib}
		}
		
		%\section{Acknowledgment}
		%This research was partly supported by the National Key R$\&$D Program of China (Grant No. 2020AAA0108303), and by Shenzhen Science and Technology Project (Grant No. JCYJ20200109143041798) $\&$ Shenzhen Stable Supporting Program (WDZC20200820200655001) $\&$ Shenzhen Key Laboratory of next generation interactive media innovative technology (Grant No: ZDSYS20210623092001004).
		% This research was partly supported by Shenzhen Key Laboratory of next generation interactive media innovative technology (Grant No: ZDSYS20210623092001004). %We sincerely thank Dr. Zhang Yan for his assistance in this work.
		
		%-------------------------------------------------------------------------
		% \newpage
		\clearpage
		\section*{Appendix}
		
		\section*{A-1. Additional Details of FineDance Dataset}
		
		FineDance contains 22 dance genres, which fall into 5 coarse categories. %The duration of different dance styles is shown in Table \ref{tab:GenreDuration}. 
		In order to make the dataset include more total time and more genres, we follow the natural distribution to collect dances. We did not deliberately pursue the balance of different dance genres, such as slightly reducing the duration of popular dances (dances with unpopular genres are difficult to collect a lot). On the contrary, FineDance aims to cover a wider range of genres, and we contend that this collection method aligns more closely with the real-world distribution of dance genres. 
		
%		\begin{table}[th]
%			\resizebox{0.5\textwidth}{!}{
%				\begin{tabular}{|l|l|c|l|l|c|}
%					\hline%\toprule [1pt] 
%					Class &  Genre &   Duration &  Class &  Genre  &  Duration \\
%					\hline	%\noalign{\smallskip}\hline\noalign{\smallskip}
%					\multirow{6}*{Street} &Locking
%					&0:07:24 & \multirow{6}*{Ballroom} &Rumba &0:53:48\\
%					~ &Breaking & 0:14:43 & ~ &Samba &0:21:44\\
%					~ &Popping & 2:22:46 & ~ &Waltz &0:18:06\\
%					~ &Urban & 0:18:46 & ~ &Tango &0:05:04\\
%					~ &Hiphop & 1:52:12 & ~ &Cha-cha &0:22:18\\
%					~ & Jazz &0:43:26 & ~ &Cowboy &0:15:56\\
%					\hline
%					\multirow{3}*{Folk} &Tai &0:51:02 & \multirow{3}*{Mix} &korean &1:11:23\\
%					~ &Uighur &0:39:29 & ~ &Choreography &0:11:47\\
%					~ &Hmong &0:27:28 & ~ &Chinese &0:09:48\\
%					
%					\hline
%					%           \multirow{4}*{Classic} &Hantang &1:09:08 & \multirow{4}*{\multicolumn{3}{*}{1}} \\
%					% ~ &Shenyun &1:40:18 &~ \\
					%           ~ &Kun & 0:02:59 &~ \\
					%           ~ &Dunhuang &0:17:07 &~ \\
					
%					\multirow{4}*{Classic} &Hantang &1:09:08 & \multirow{3}*{ } &  & \\
%					~ &Shenyun &1:40:18 & ~ & & \\
%					~ &Kun &0:02:59 & ~ &  & \\
%					~ &Dunhuang &0:17:07 & ~ &  & \\
					
%					\hline%\bottomrule [1pt] 
%					% \noalign{\smallskip}
%				\end{tabular}
%			}
%			\caption{Duration of different genres.}\vspace{-3mm}
%			\label{tab:GenreDuration}   
%		\end{table}

		\section*{A-2. Additional Details of FDGN}
		The body/hand expert nets are designed based on the diffusion model\cite{ho2020denoising}.
		% Diffusion networks usually consist of forward and reverse processes respectively. The diffusion process is to continuously add noise until the data obeys the Gaussian distribution; the revere diffusion process is to sample a noise from the Gaussian distribution, and then iteratively generate the data.
		A diffusion process gradually perturbs the data towards random noise, while a deep neural network learns to denoise.
		% Unlike the VAE \cite{VAE} or Flow-based models \cite{rezende2015variational}, {\color{yellow}the diffusion model does not compress features????}, so the generated results are expressive and diverse.
		Compared with VAE\cite{VAE} or Flow-based models\cite{rezende2015variational}, diffusion models do not encode features into high-dimensional embedding spaces, thus generating expressive and diverse results. Compared with GAN\cite{GAN} models, diffusion models are easier to converge.
		
		Given the music temporal feature ${\bar{X}}^t$, the body expert diffusion model of FDGN can generate corresponding body dance fragment ${\hat{Yb}}^t$, where $t$ represents the $t-th$ fragment. For readability, we do not indicate the superscript $t$ in this document. Given a body dance fragment $Yb_0$ with 4 seconds, where $Yb_0\sim q(Yb_0)$ is the ground truth dance data. The diffusion process of 3DGNet is to add Gaussian noise in $S$ steps, and gradually change $Yb_0$ to $Yb_1,Yb_2,\cdots,Yb_S$. When $S$ is large enough, $Yb_S\sim \mathcal{N}(0,I)$. The diffusion process can be formulated by a Markov chain:

		\begin{equation}
			\begin{split}
				\begin{array}{lr}
					q\left(Yb_{1:S}| Yb_{0}\right) =\displaystyle \prod ^{S}_{s=1} q\left(  Yb_{s}| Yb_{s-1}\right), \vspace{1ex}  \\
					q\left(Yb_{s}| Yb_{s-1}\right) =\mathcal{N}(Yb_{s};\sqrt{1-\beta_{s}}Yb_{s-1};\beta_{s}I), 
				\end{array}
			\end{split}
		\end{equation}
		where $\beta_{s}$ is the variance of the Gaussian noise added at diffusion step $s$.
		
		We follow \cite{ho2020denoising, zhang2022motiondiffuse} to simplify this multi-step diffusion process into one step, which can be formulated as:
		\begin{equation}
			\begin{aligned}
				Yb_s &= \sqrt{\alpha_s}Yb_{s-1}+\sqrt{1-\alpha_s}\boldsymbol\epsilon_{s-1} \\
				&= \sqrt{\alpha_s\alpha_{s-1}}Yb_{s-2}+\sqrt{1-\alpha_s\alpha_{s-1}}\bar{\boldsymbol\epsilon}_{s-2} \\
				&= \cdots \\
				&= \sqrt{\bar{\alpha}_s}Yb_{0} + \sqrt{1-\bar{\alpha}_s}\boldsymbol\epsilon, \\
				\alpha_{s} &= 1-\beta_{s},\bar{\alpha}_{s}= \displaystyle \prod ^{s}_{i=1}\alpha_s,
			\end{aligned}
		\end{equation}
		%\{X^t\}_{t=1}^N
		where $\boldsymbol\epsilon_{s-1}$ is the Gaussian noise added at diffusion step $s$, $\bar{\epsilon}_{s-2}$ merges two Gaussian noises, and $\boldsymbol\epsilon$ merges $s$ Gaussian noises, $\boldsymbol\epsilon \sim \mathcal{N}(0,I)$. Then the diffusion process can be formulated as:
		\begin{equation}
			\begin{split}
				\begin{array}{lr}
					q\left(Yb_{s}| Yb_{0}\right) =\mathcal{N}(Yb_{s};\sqrt{1-\bar{\alpha}_{s}}Yb_{0};\bar{\alpha}_{s}I).
				\end{array}
			\end{split}
		\end{equation}
		%将数据集中的GroundTruth切分好的舞蹈片段表示为$Yb$，
		%3DGNet的前向过程是从真实舞蹈
		
		Conditioned on the input music clip temporal feature $\bar{X}$, the reverse diffusion process of Body Expert diffusion Network $BEN$ is to estimate $\hat{Yb}_{0}$ from $Yb_{S}$. Instead of predicting $\boldsymbol\epsilon_{s}$ at each diffusion step, we follow MDM and directly estimate $\hat{Yb}_{0}$ as, i.e., $\hat{Yb}_{0}=BEN({Yb}_s,s,\bar{X})$. The training objective can be formulated as:
		\begin{equation}
			\mathcal{L}_d=\mathrm{E}_{s \in[1, S], Yb_0 \sim q\left(Yb_0\right)}[\left\|{Yb_0}-{BEN}\left(Yb_s, s, \bar{X}\right)\right\|_2^2]
		\end{equation}
		
		The structure of hand expert diffusion network is similar to that of body expert diffusion network, except that we concat the output of body expert net and the music features together as the condition of hand expert net to generate flexible and natural hand motions coordinate with body.
		
		The generation process of FDGN is also the reverse diffusion process. First, $Yb_S$ and $Yh_S$ are sampled from the Gaussian distribution, and then the dance data is iteratively restored. By sampling different noises, diverse dance fragments can be generated.

		\section*{A-3. Potential Applications of FineDance Dataset}
		% 介绍数据集在其他领域的应用，如人体重建、body/hands estimation
		Apart from generating dances from music, FineDance dataset can also be used in the following tasks:
		
		\noindent\textbf{Motion Prior.}
		%There are some patterns in human motion, 
		Ignoring Motion Prior and relying solely on position and velocity to estimate motion may result in unnatural movements. Therefore, some works such as Mposer\cite{Mposer} and Vposer\cite{SMPLX} began to study Motion Prior, LEMO\cite{lemo} employs Motion Prior to solve the occlusion problem in human reconstruction. Ma et al.\cite{PretrainedMotion} pre-train on multiple datasets to learn motion prior, achieving better results in tasks such as motion generation and prediction.
		FineDance has 14.6 hours of accurate annotation. The flexible and natural dance movements are different from existing datasets, allowing Motion Prior to obtain richer prior knowledge of human motion.
		
		\noindent\textbf{Generating Music from Dances.}
		Music is naturally connected with dance. There has been a lot of work on generating dance from music, but its inverse task has not been fully explored.
		Since the music and dance in the FineDance dataset are strictly aligned, it also supports the task of generating music from dance. 
		
		%\noindent\textbf{3D Pose Estimation.} This task requires the algorithms to automatically estimate the 3D position of human joints from RGB images or videos\cite{lu2018review,luvizon20182d,pavllo20193d}. Based on the accurate 3D joints information and multi-view videos, our dataset can also be used for 3D body/hand pose estimation.
		
		%\noindent\textbf{Full body reconstruction.} The goal of 3D full-body reconstruction is to recover the mesh of the human body and hands from images or videos. With the emergence of SMPL-X\cite{SMPLX}, a parametric model based on human anatomy and dynamics that is capable of expressing 3D human full-body pose and shape,  researchers endeavor to achieve  full-body reconstruction by estimating SMPL-X parameters. However, due to the complexity and high cost of hand motion capture, there are few datasets that contain both body and hand annotations, which limits the development of full-body reconstruction algorithms. For example, FuRPE\cite{FuRPE} and Dope\cite{Dope} use pseudo-label methods to alleviate the problem of insufficient training data.
		
		Our FineDance dataset contains accurate body and hand parameters of SMPL-X captured by the motion capture system, as well as multi-view RGB videos, which can provide support for full-body reconstruction.
		And since FineDance contains a large number of flexible dance gestures, which are real and natural movements, but rare in other datasets. Therefore, training on the FineDance dataset may improve the generalization of the model.

		\section*{A-4. Additional Qualitative Results}
		% 视频截图、与定性分析
		
		We show the additional qualitative results of our work in the following five aspects. We strongly encourage you to watch the video for more details.

		\noindent\textbf{FineDance Dataset Show.} Due to the limitation of file size, we show 8 genres dance of our dataset, which are Korea, Jazz, Breaking, Locking, Hmong, Uighur, DunHuang, ShenYuan. The result is shown in our video and Figure \ref{fig:dataset_show}.
		
		\noindent\textbf{Ablation experiment of FDGN.} 
		In order to verify whether FDGN can generate more flexible and natural hand movements, we replaced FDGN with a diffusion model without expert nets and refined nets, and only used the diffusion model to simultaneously generate body and hand dances. We named this experiment as FineNet (w/o FDGN). As can be seen from Figure \ref{fig:hand_compare}, the second row shows unnatural hand gestures with excessive reverse distortion, the third row of hand movements is tedious, and the first row shows FineNet with FDGN can generate flexible and natural hand movements.
		
		\noindent\textbf{Muti-genres dance generation.} We use FineNet to generate dance fragments with different genres.
		%generate 8 dance fragments of 8 different dance genres to show the diverse genre generation ability of our method.
		The results are shown in our video and Figure \ref{fig:multi-gen}. Comparing the different rows in Figure \ref{fig:multi-gen}, we can see that our method can generate dances of different genres, which vary widely. Comparing the same rows in Figure \ref{fig:dataset_show} and Figure \ref{fig:multi-gen}, we can see that the characteristics of the generated dances of different genres are consistent with the real dances, while the generated dances have great novelty.
		
		\noindent\textbf{Compare With The SOTAs.} Based on the FineDance dataset, we use FineNet and other methods to generate 20 second dance.  For DeepDance and DanceRevolution, the generated dances are repetitive and meaningless. ChoreoMaster directly retrieves and stitches dances from the ground truth dance fragments. Although the dances are real and effective, the genre accuracy is bad. As shown in the last two pictures in the third row of Figure \ref{fig:com_sotas}, the classical dance fragment appears in street dance. It can be observed that our method is more comprehensive.
		
		\noindent\textbf{Generate Diverse Dances Form the Same Music. } We use FineNet to generate 6 dance fragments from only one music clip. The results are shown in our video and Figure \ref{fig:MM1} \ref{fig:MM2} \ref{fig:MM3}. The results show that the 6 dance fragments are completely different, which demonstrates the excellent multimodality of our method. 
		
		\newpage
		
		\begin{figure*}[t]
			\centering
			\includegraphics[width=1\textwidth]{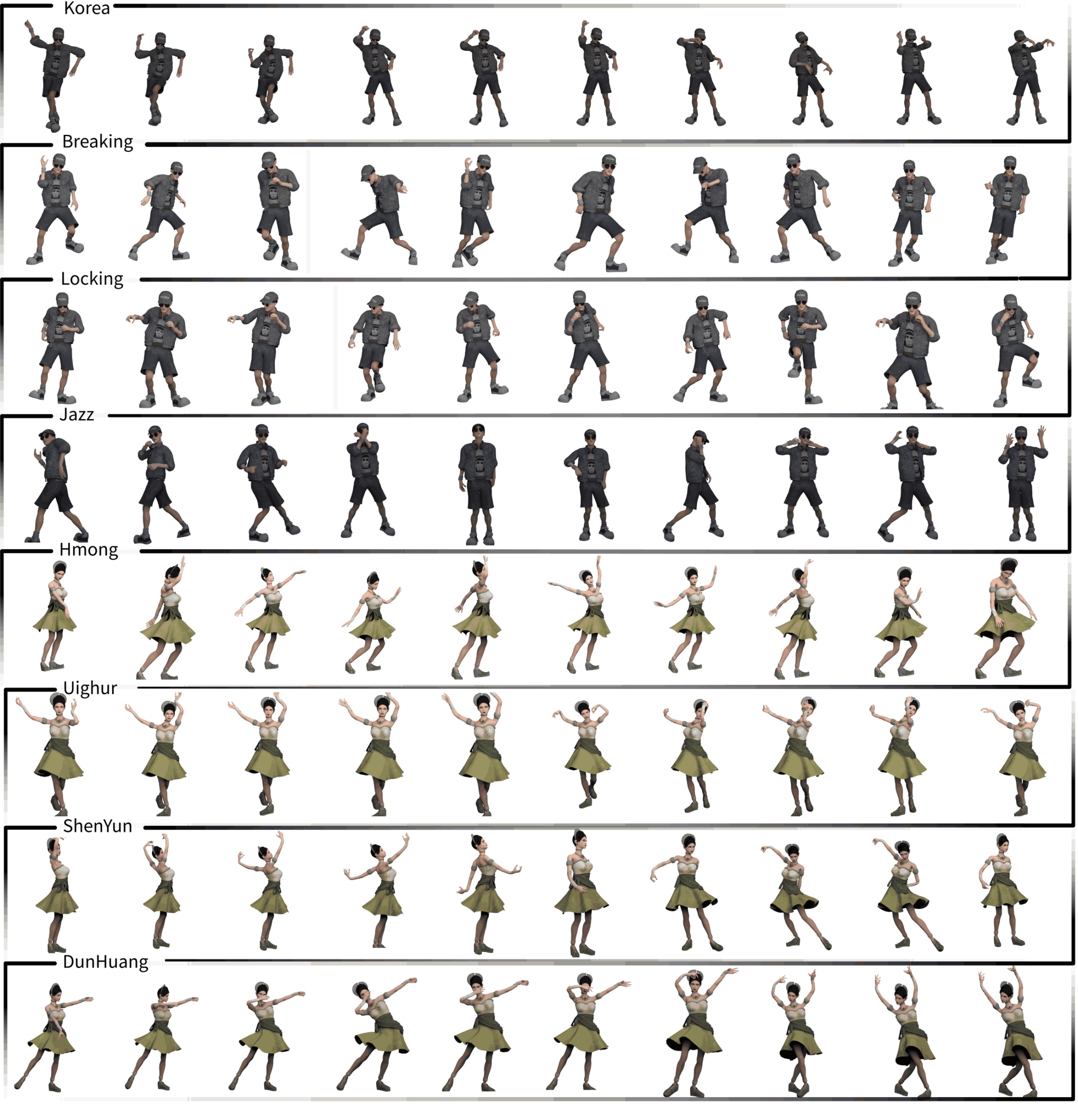}
			\caption{FineDance Dataset show.}\vspace{-3mm}
			\label{fig:dataset_show}
		\end{figure*}
		
		\begin{figure*}[t]
			\centering
			\includegraphics[width=1\textwidth]{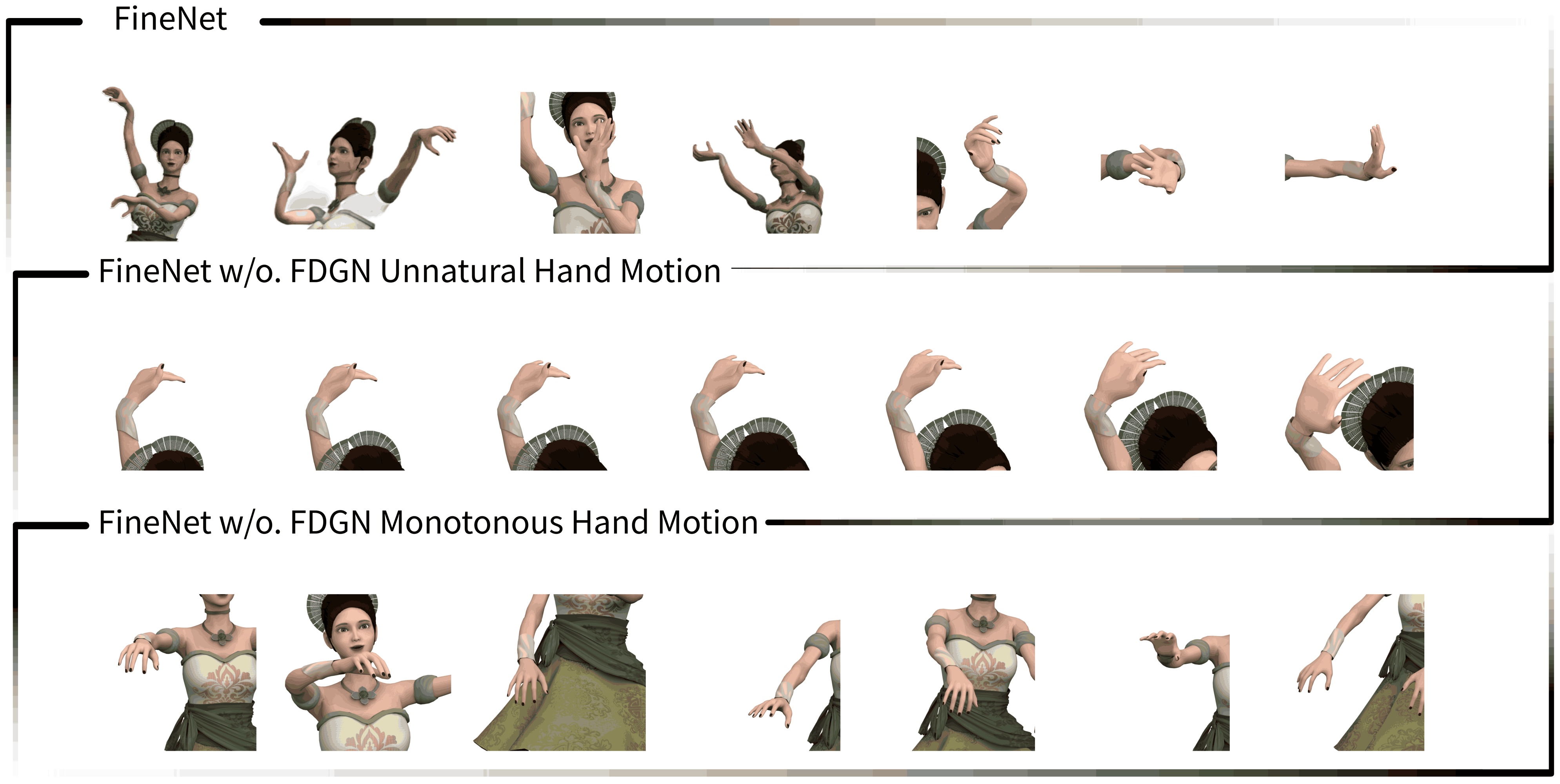}
			\caption{Ablation experiment of FDGN.}\vspace{-3mm}
			\label{fig:hand_compare}
		\end{figure*}
		
		\begin{figure*}[t]
			\centering
			\includegraphics[width=1\textwidth]{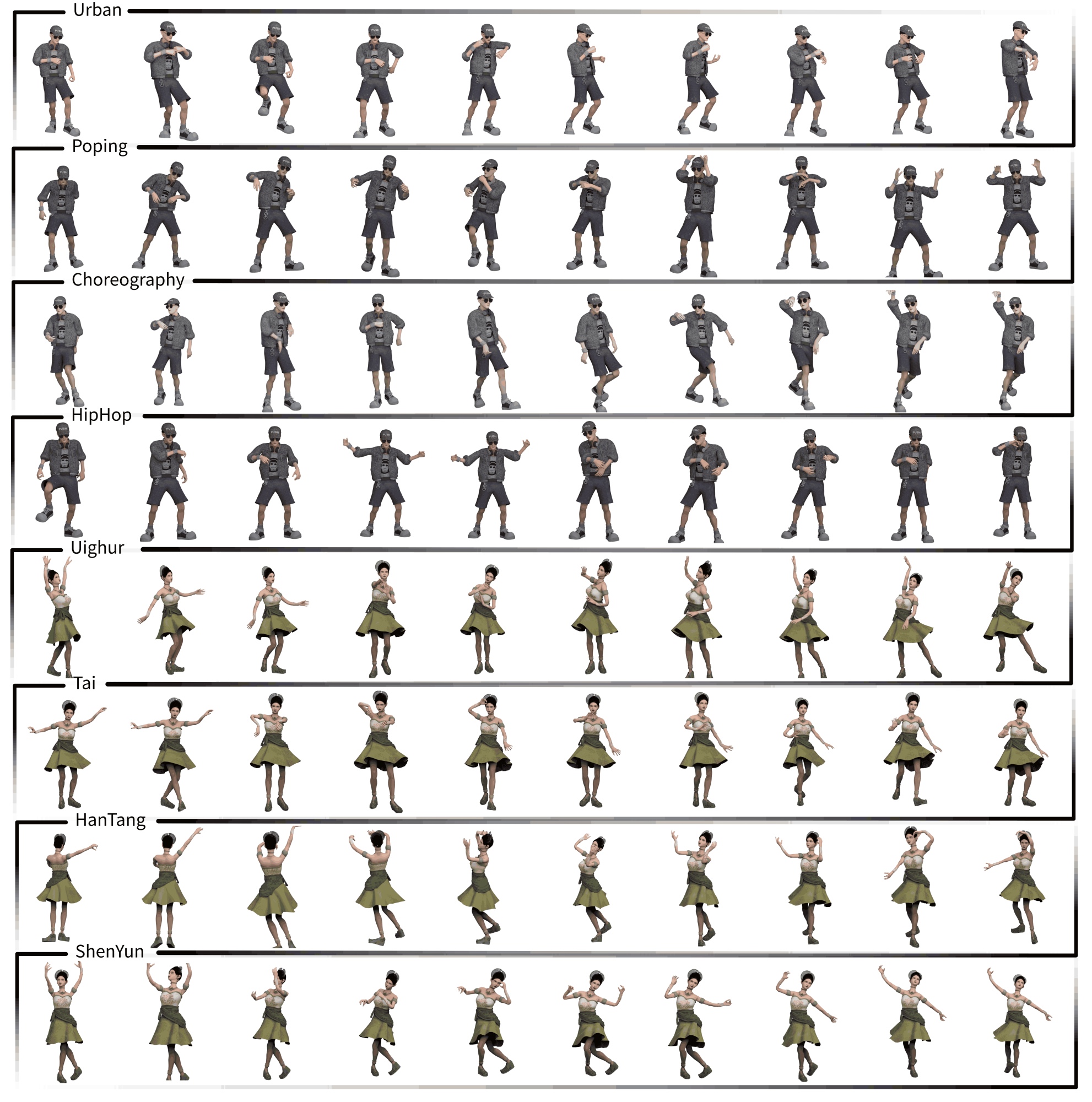}
			\caption{Generated multi-genre dance.}\vspace{-3mm}
			\label{fig:multi-gen}
		\end{figure*}
		
		\begin{figure*}[t]
			\centering
			\includegraphics[width=1\textwidth]{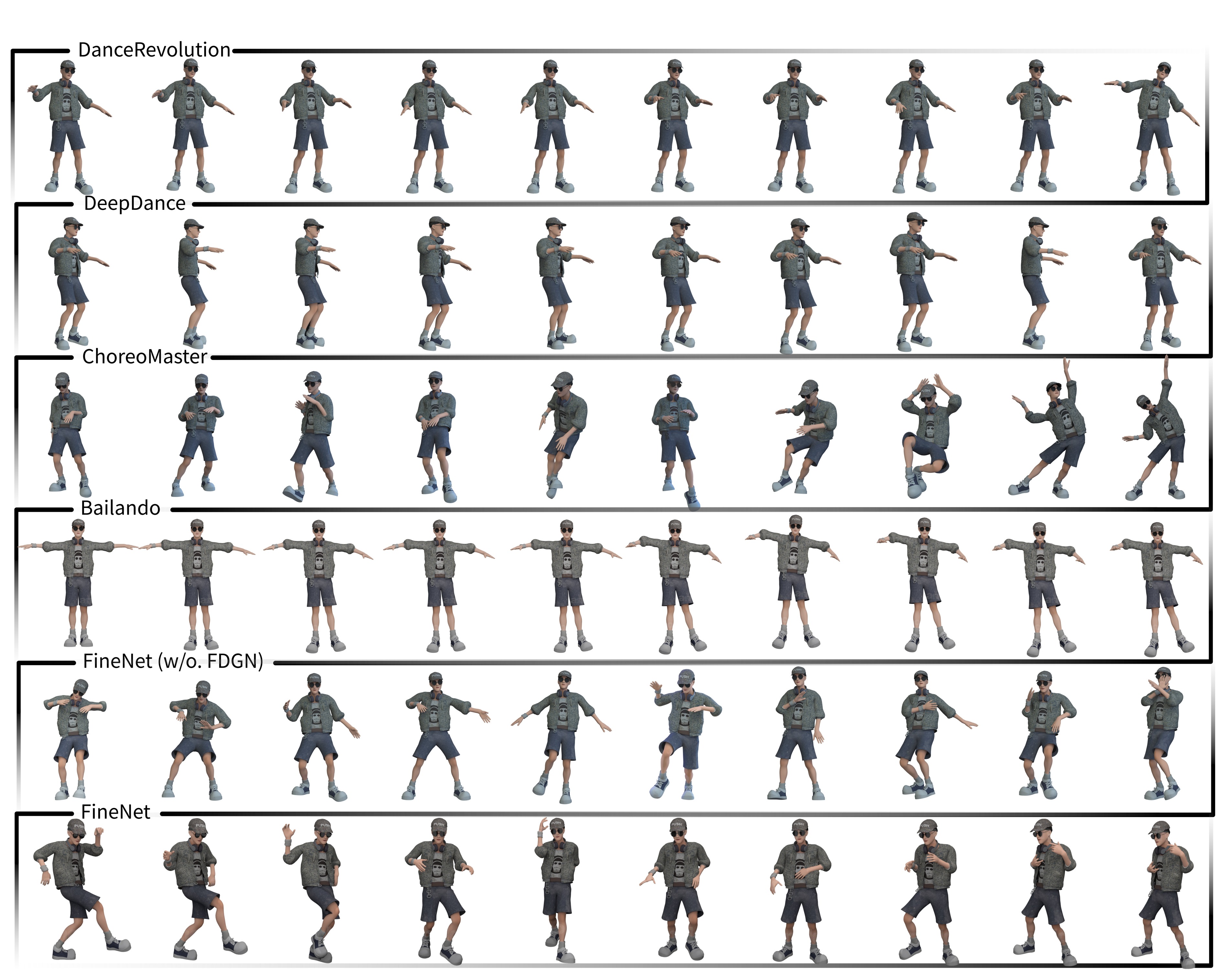}
			\caption{Compare with the SOTAs.}\vspace{-3mm}
			\label{fig:com_sotas}
		\end{figure*}

		\begin{figure*}[t]
			\centering
			\includegraphics[width=1\textwidth]{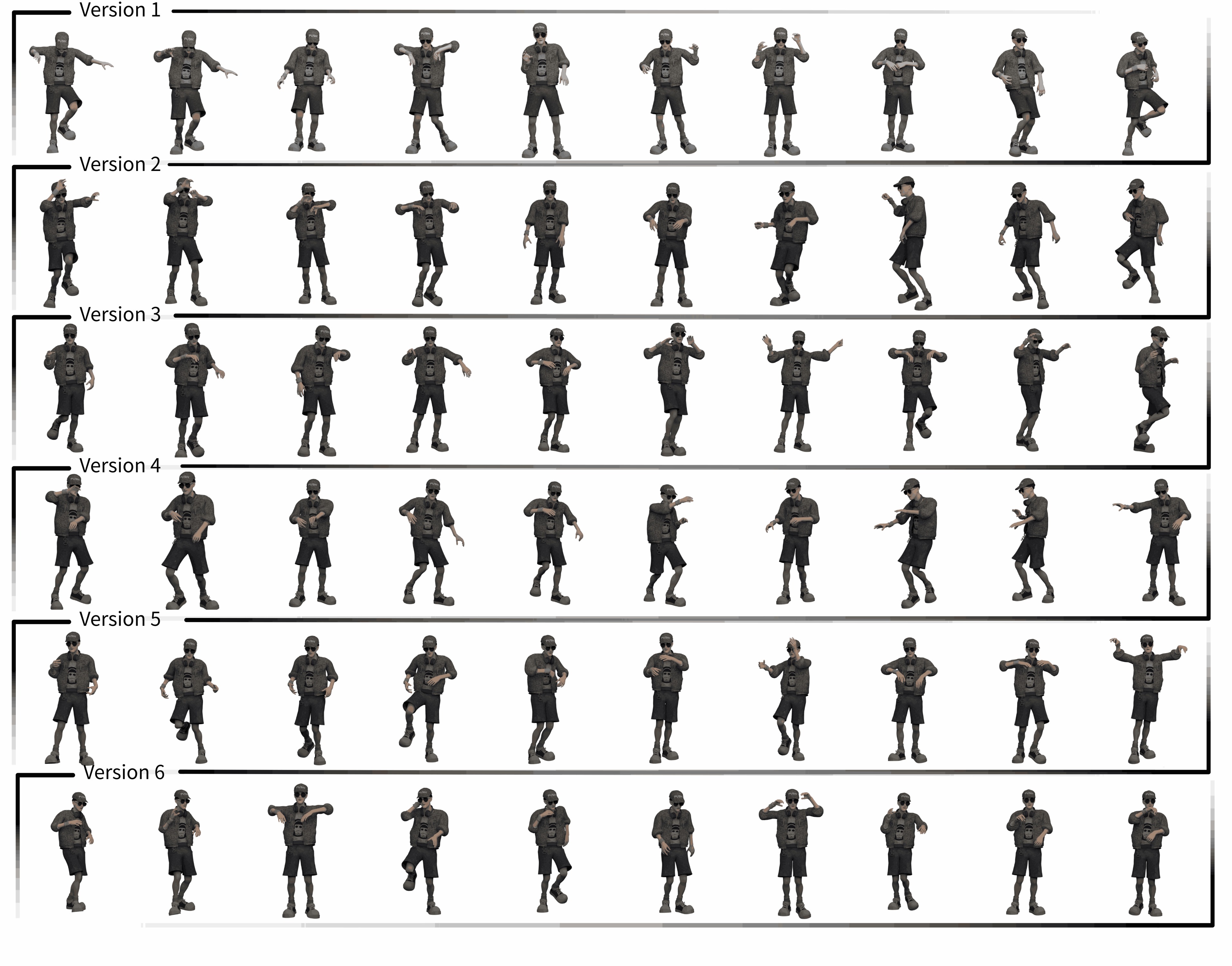}
			\caption{Generated diverse dances form the same music. By music \textit{Goodbye Monica}.}
			\label{fig:MM1}
		\end{figure*}
		
		\begin{figure*}[t]
			\centering
			\includegraphics[width=1\textwidth]{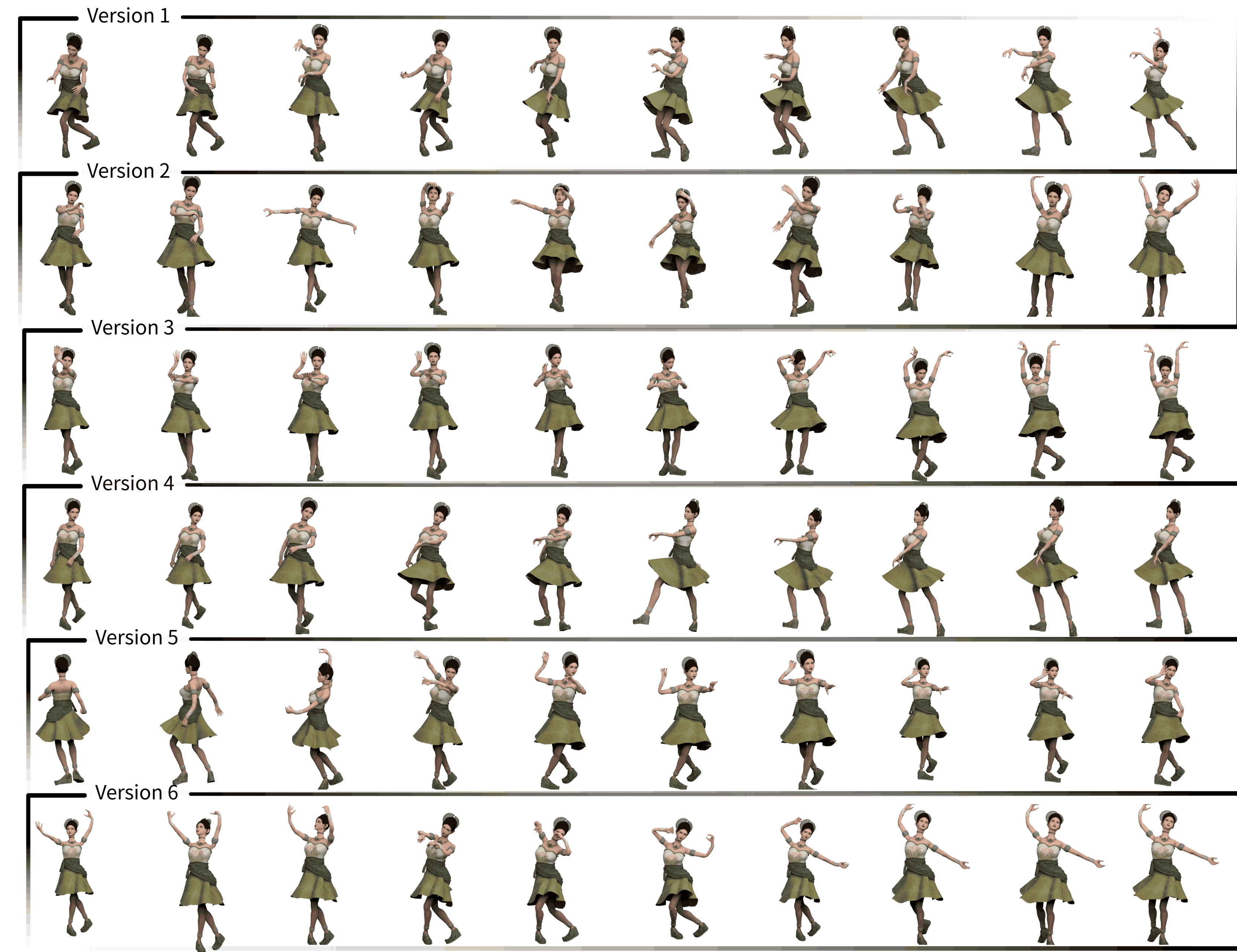}
			\caption{Generated diverse dances form the same music. By music \textit{Dreams Come True}. }
			\label{fig:MM2}
		\end{figure*}
		
		\begin{figure}[t]
			\hsize=\textwidth  
			\centering
			\includegraphics[width=1\textwidth]{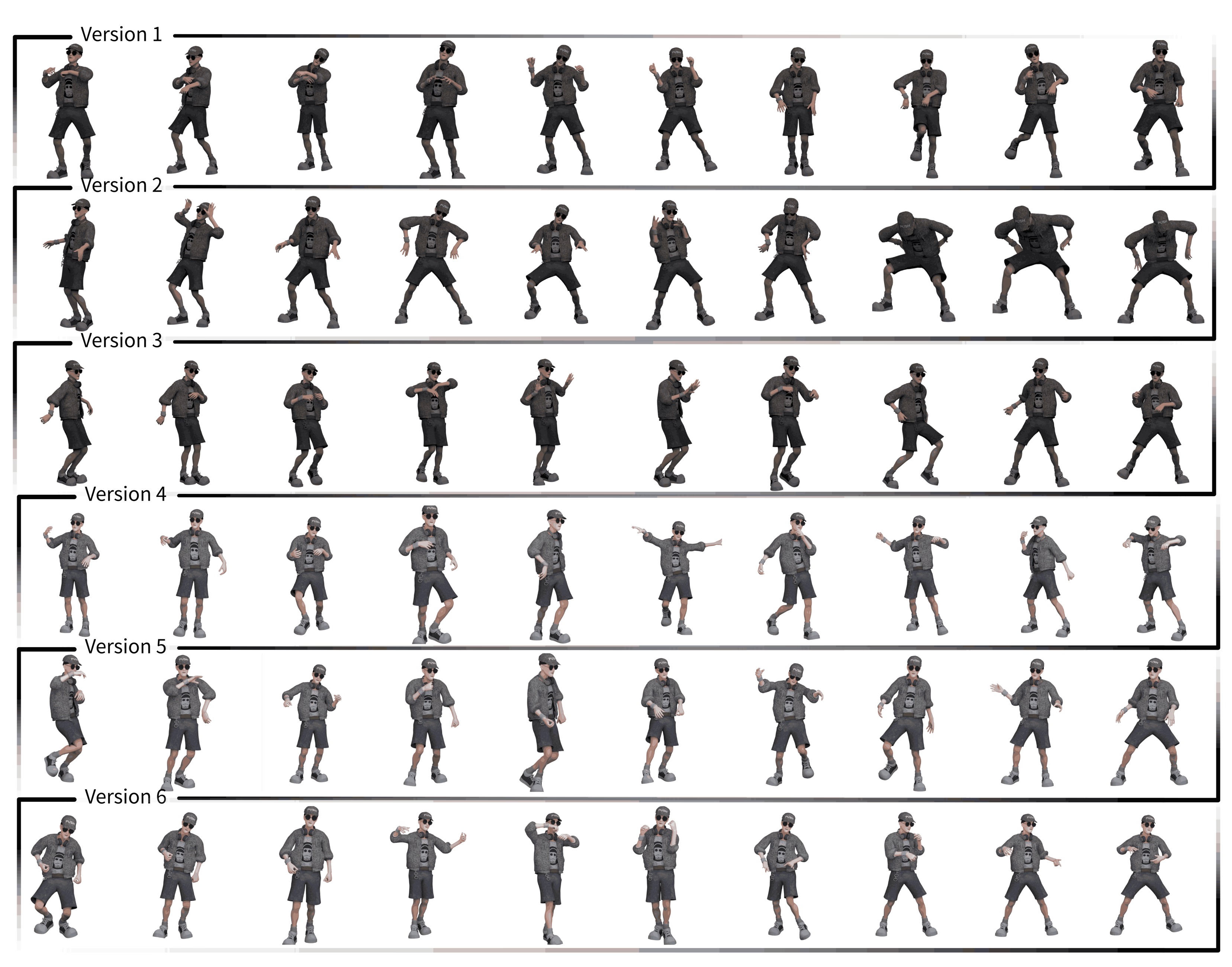}
			\caption{Generated diverse dances form The same music. By music \textit{Just Begun}.}
			\label{fig:MM3}
		\end{figure}

	\end{document}